\documentclass[titlepage]{csetr}

\usepackage{graphicx}

\usepackage{fancyvrb}

\usepackage{setspace}
\usepackage{geometry}
\usepackage{float}
\usepackage{setspace,supertabular}
\usepackage{longtable}
\usepackage[usenames]{color}
\usepackage{verbatim}
\usepackage{subfigure}
\usepackage{amsmath}
\usepackage[table]{xcolor}

\usepackage{lscape}
\usepackage{changepage}
\usepackage{tabularx}

\renewcommand{\and}{\hspace{.5cm}}

\trnumber{201333}
\trdate{November 2013}

\title{%
 Big Data and Cross-Document Coreference Resolution: Current State and Future Opportunities
}
\author{%
        Seyed-Mehdi-Reza Beheshti \\
        Srikumar Venugopal \\
        Seung Hwan Ryu \\
        Boualem Benatallah \\
        Wei Wang \\\\
  $\, $University of New South Wales\\ Sydney 2052, Australia \\%
  \email{\{sbeheshti,srikumarv,seungr,boualem,weiw\}@cse.unsw.edu.au}\\ \\
}

\date{}
\begin{document}
\maketitle

\begin{abstract}

Information Extraction (IE) is the task of automatically extracting structured information from unstructured/semi-structured machine-readable documents. Among various IE tasks, extracting actionable intelligence from ever-increasing amount of data depends critically upon Cross-Document Coreference Resolution (CDCR) - the task of identifying entity mentions across multiple documents that refer to the same underlying entity. Recently, document datasets of the order of peta-/tera-bytes has raised many challenges for performing effective CDCR such as scaling to large numbers of mentions and limited representational power. The problem of analysing such datasets is called ``big data".
The aim of this paper is to provide readers with an understanding of the central concepts, subtasks, and the current state-of-the-art in CDCR process. We provide assessment of existing tools/techniques for CDCR subtasks and highlight big data challenges in each of them to help readers identify important and outstanding issues for further investigation. Finally, we provide concluding remarks and discuss possible directions for future work.

\end{abstract}

\section{Introduction}
\label{chap1}

The majority of the digital information produced globally is present in the form of web pages, text documents, news articles, emails, and presentations expressed in natural language text. Collectively, such data is termed \emph{unstructured} as opposed to \emph{structured} data that is normalised and stored in a database. The domain of information extraction (IE) is concerned with identifying information in unstructured documents and using it to populate fields and records in a database~\cite{mccallum2005ie}. In most cases, this activity concerns processing human language texts by means of natural language processing (NLP)~\cite{spyns1996natural}.

Among various IE tasks, Cross-Document Coreference Resolution (CDCR)~\cite{MayfieldADEE09,Bagga1998} involves identifying equivalence classes for identifiable data elements, called entities, across multiple documents. In particular, CDCR is of critical importance for data quality and is fundamental for high-level information extraction and data integration, including semantic search, question answering, and knowledge base construction.

Traditional approaches to CDCR~\cite{MayfieldADEE09,wellner2004integrated} derive features from the context surrounding the appearance of an entity in a document (also called a ``mention") and then apply clustering algorithms that can group similar or related entities across all documents. As we will soon discuss, these approaches aim for being exhaustive and grow exponentially in time with the increase in the number of documents.


Recently, a new stream of CDCR research~\cite{elsayed2008pairwise,pantel2009web,sarmento2009approach,singh2011large,kolb2012dedoop} has focused on meeting the challenges of scaling CDCR techniques to deal with document collections sized of the order of tera-bytes and above. Popularly, dealing with such large-scale datasets has been termed as the ``big data" problem. In this context, CDCR tasks may face various drawbacks including difficulties in clustering and grouping large numbers of entities and mentions across large datasets. Therefore, CDCR techniques need to be overhauled to meet such challenges.


To address these challenges, researchers have studied methods to scale CDCR subtasks such as computing similarity between pairs of entity mentions~\cite{ng2010supervised,wick2009entity}, or even to replace pairwise approaches with  more expressive and scalable alternatives~\cite{fastCoreference1,wellner2004integrated}. Recent publications~\cite{elsayed2008pairwise,pantel2009web,sarmento2009approach,singh2011large,kolb2012dedoop} have reported on the usage of parallel and distributed architectures such as Apache Hadoop~\cite{Hadoop,MapReduce} for supporting data-intensive applications which can be used to build scalable algorithms for pattern analysis and data mining.

Although these techniques represent the first steps to meeting big data challenges, CDCR tasks face various drawbacks in achieving a high quality coreference result (effectiveness) and performing the coreference resolution as fast as possible (efficiency) on large datasets. The aim of this paper is to provide readers with an understanding of the central concepts, subtasks, and the current state-of-the-art in CDCR process. We assess existing tools/techniques for CDCR subtasks and highlight big data challenges in each of them to help readers identify important and outstanding issues for further investigation. Finally, we provide concluding remarks and discuss possible directions for future work.

The remainder of this document is organized as follows. In Section~\ref{sec2}, we introduce the CDCR process and its sub-tasks in detail. Sections~\ref{chap3} and~\ref{chap4} discuss the state-of-the-art in entity identification and entity classification. Section~\ref{chap5} discusses the challenges brought about by big data in CDCR. Section~\ref{sec6} present the state-of-the-art tools and techniques for CDCR. Section~\ref{chap6} presents our conclusions and a roadmap for the future. Finally, in the Appendix we discuss our experience in implementing a MapReduce-based CDCR software prototype to address challenges discussed in the paper.


\section{CDCR Process and Evaluation Framework}
\label{sec2}

\subsection{Background and Preliminaries}
\label{chap1}

\begin{figure}
\centering
\includegraphics[width=0.8\textwidth]{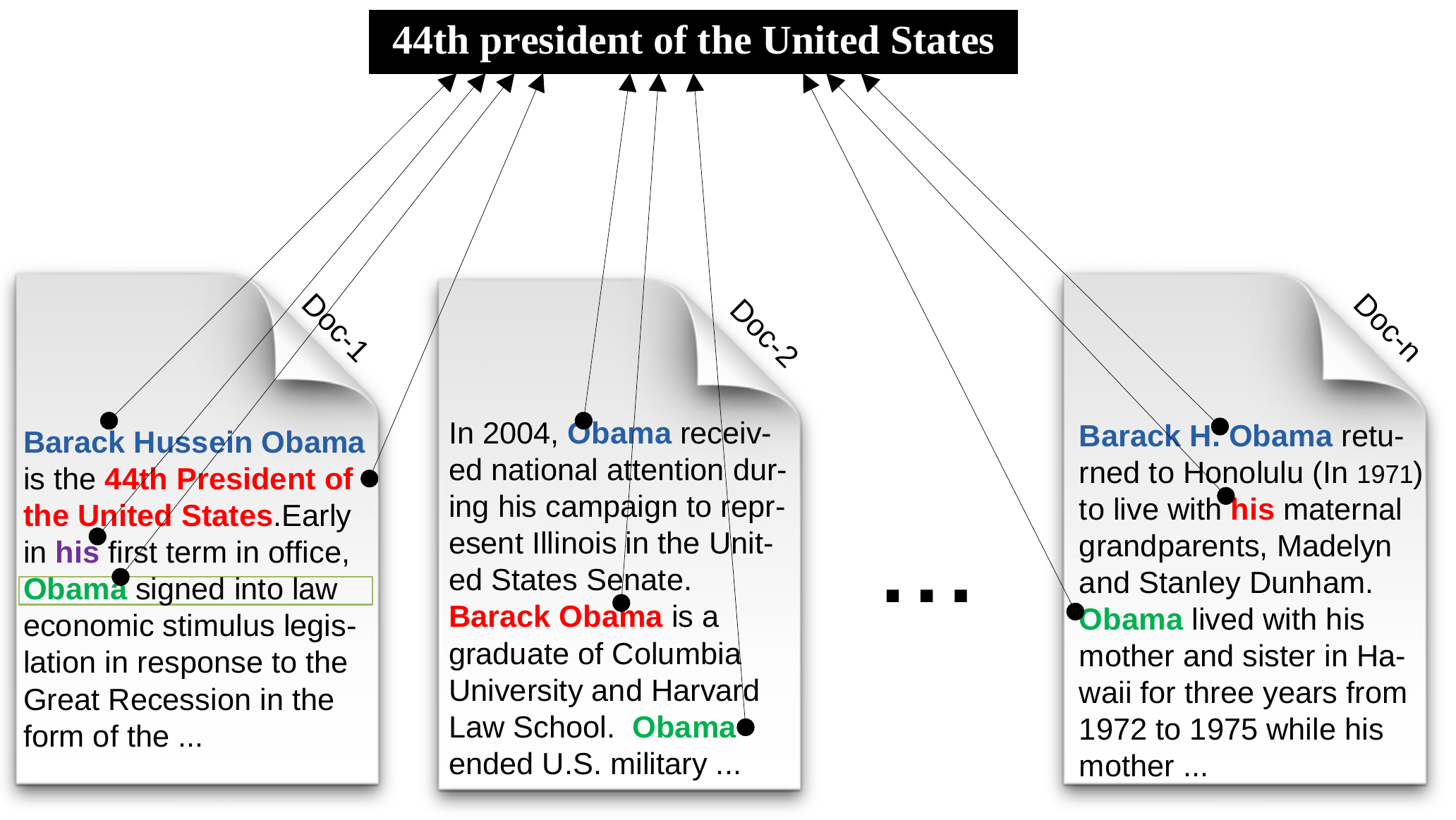}
\caption{An \emph{entity} (i.e. the person who serves as the 44th president of the United States) and its \emph{entity mentions}, i.e. its true coreference resolutions. }
\label{fig:CDCExample}
\end{figure}

CDCR approaches provide techniques for the identification of entity mentions in different documents that refer to the same underlying entity. In this context, an \emph{entity} is a real-world person, place, organization, or object, such as the person who serves as the 44th president of the United States and an \emph{entity mention} is a string which refers to such an entity, such as ``Barack Hussein Obama", ``Senator Obama" or ``President Obama". Figure~\ref{fig:CDCExample} illustrates a sample example of person name mentions from different documents and their coreference resolutions. Given a collection of mentions of entities extracted from millions of documents, CDCR involves various subtasks, from extracting entities and mentions to clustering the mentions. The overall objective is to cluster mentions such that mentions referring to the same entity are in the same cluster and no other entities are included~\cite{singh2011large}. Mentions referring to the same entity are termed ``co-referent".

The current approach to cross-document (named) entity coreference resolution consists of two primary tasks~\cite{MayfieldADEE09,fastCoreference1,finin2009using,rao2010streaming,dozier2004cross}: entity identification and classification. Entity identification is the process of finding mentions of the entities of interest in documents and tie together those that are coreferent, while entity classification task involves deriving a classification and/or clustering technique that will separate data into categories, or classes, characterized by a distinct set of features. We discuss each in depth in the following subsections.

\subsubsection{Entity Identification}




Named-entity recognition~\cite{NER.evaluation.2009,NERComparison} (NER), also known as entity identification~\cite{nadeau2007survey} and entity extraction~\cite{chen1998named,ah2009clique}, refers to techniques that are used to locate and classify atomic elements in text into predefined categories such as the names of persons, organizations, locations, expressions of times, quantities, monetary values, percentages, etc. There are numerous approaches and systems available for performing NER. For example, for traditional named entity recognition (NER), the most popular publicly available systems are: OpenNLP NameFinder\footnote{http://opennlp.apache.org/}, Illinois NER system\footnote{http://cogcomp.cs.illinois.edu/demo/ner/?id=8}, Stanford NER system\footnote{http://www-nlp.stanford.edu/software/CRF-NER.shtml}, and Lingpipe NER system\footnote{http://alias-i.com/lingpipe/demos/tutorial/ne/read-me.html}.

Various steps are considered in this task including: (i)~Format Analysis, in which document formats are analysed for  formatting information in addition to textual content; (ii)~Tokeniser, where text is segmented into tokens, e.g., words, numbers, and punctuation; (iii)~Gazetteer, where the type and scope of the information is categorized; and (iv)~Grammar, where linguistic grammar-based techniques as well as statistical models are used to extract more entities. The output of entity identification task will be a set of named entities, extracted from a set of documents. Numerous approaches, techniques, and tools to extracting entities from individual documents have been described in the literature and will be discussed in depth in the next section.


\subsubsection{Entity Classification}

Entity classification task involves deriving a classification and/or clustering technique that will separate data into categories, or classes, characterized by a distinct set of features. To achieve this, extracted entities and mentions (from the entity identification task) are assigned a metric based on the likeness of their meaning or semantic content.

\begin{figure}
\centering
  \includegraphics[scale=0.61]{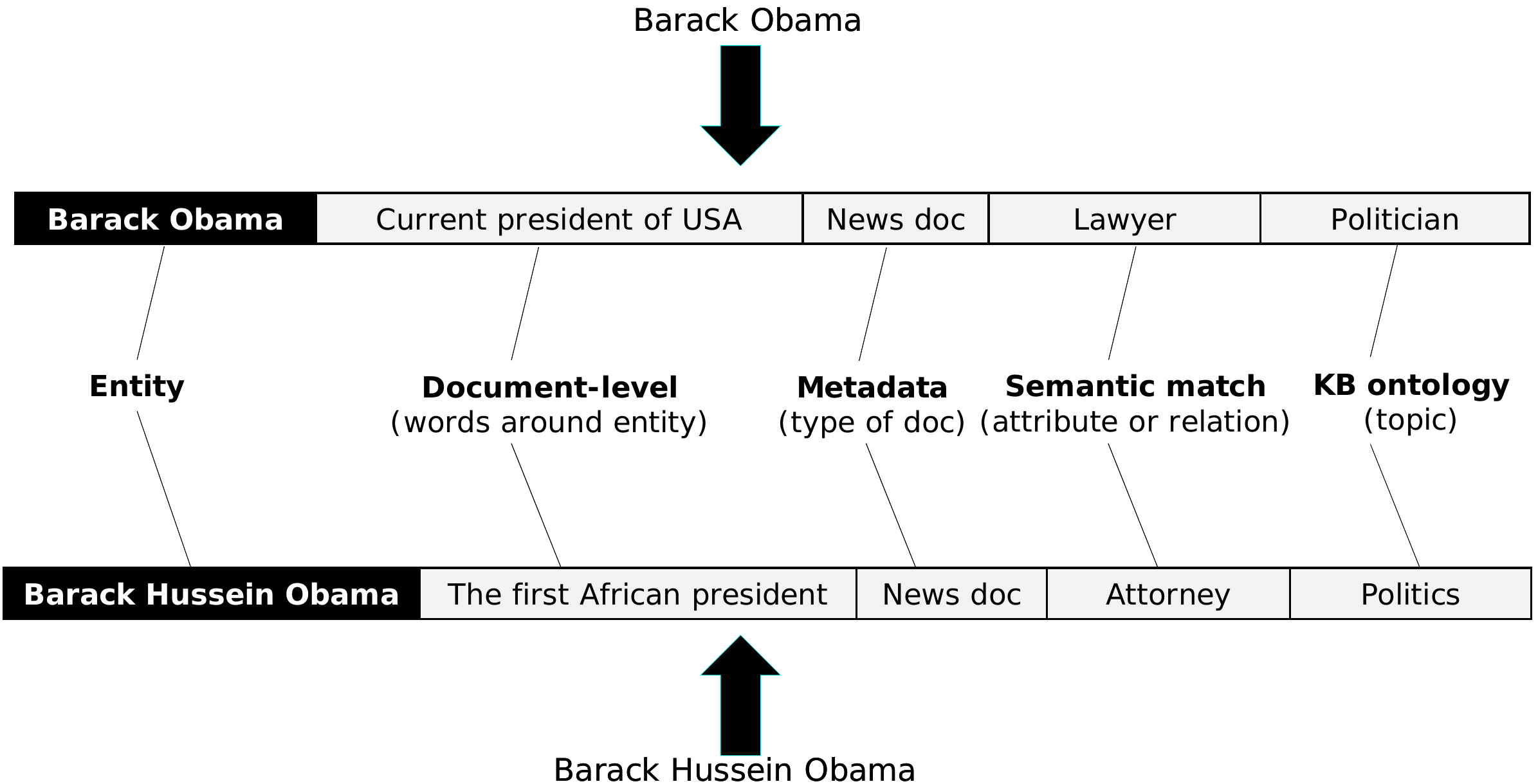}\\
  \caption{A simple example for calculating various featurization classes for the pair of entities (`Barack Obama' , `Barack Hussein Obama').}\label{featurization}
\end{figure}

Various machine learning techniques have modeled the problem of entity coreference as a collection of decisions between mention pairs~\cite{fastCoreference1}. Prior to entity pairing, various \emph{features} may be extracted to annotate entities and their mentions. Figure~\ref{featurization} illustrates a simple example for calculating various featurization classes for the pair of mentions \{`Barack Obama' , `Barack Hussein Obama'\}. As illustrated in the figure, these classes can be defined for entities, words around the entities (document level), and meta-data about the documents such as their type. Then, the similarity scores for a pair of entities are computed  using appropriate similarity functions for each type of feature (e.g., character-, document-, or metadata-level features).

The next step in entity classification task is to determine whether pairs of entities are co-referent or not. For example, in the sentence ``Mary said she would help me", \emph{she} and \emph{Mary} most likely refer to the same person or group, in which case they are co-referent. Several filtering steps can be applied to entity pairs to eliminate those pairs that have little chance of being deemed co-referent.  Various supervised and/or unsupervised classification/clustering techniques (over a set of training examples) can be used to classify related entities. For example, generative classifiers (e.g., Hidden Markov model), discriminative classifiers (e.g., Support Vector Machine (SVM) or maximum entropy model), or decision tree techniques can be used to separate a set of featurized paired entities into two possible classes - coreferent or not-coreferent.

\section{Entity Identification: State-of-the-Art}
\label{chap3}

Named Entity Recognition (NER), also known as Entity Extraction (EE), techniques can be used to locate and classify atomic elements in text into predefined categories such as the names of persons, organizations, locations, expressions of times, quantities, monetary values, and percentages. NER is a key part of information extraction system that supports robust handling of proper names essential for many applications, enables pre-processing for different classification levels, and facilitates information filtering and linking. However, performing coreference, or entity linking, as well as creating templates is not part of NER task.


A basic entity identification task can be defined as follows: \\

\noindent \emph{Let \{$t_1$, $t_2$, $t_3$, ..., $t_n$\} be a sequence of entity types denoted by $T$ and let \{$w_1$, $w_2$, $w_3$, ..., $w_n$\} be a sequence of words denoted by $W$, then the identification task can be defined as `given some $W$, find the best $T$'.}\\

In particular, entity identification consists of three subtasks: entity names, temporal expressions, and number expressions, where the expressions to be annotated are `unique identifiers' of entities (organizations, persons, locations), times (dates, times), and quantities (monetary values, percentages).
Most research on entity extraction systems has been structured as taking an unannotated block of text (e.g., ``Obama was born on August 4, 1961, at Gynecological Hospital in Honolulu") and producing an annotated block of text, such as the following\footnote{In this example, the annotations have been done using so-called ENAMEX (a user defined element in the XML schema) tags that were developed for the Message Understanding Conference in the 1990s.}:

\begin{verbatim}
  <ENAMEX TYPE="PERSON">Obama</ENAMEX> was born on
  <TIMEX TYPE="DATE">August 4, 1961,</TIMEX> at
  <ENAMEX TYPE="ORGANIZATION">Gynecological Hospital</ENAMEX> in
  <ENAMEX TYPE="CITY">Honolulu</ENAMEX>.
\end{verbatim}

where, entity types such as person, organization, and city are recognized. 

%


However, NER is not just matching text strings with pre-defined lists of names. It should recognize entities not only in  contexts where category definitions are intuitively quite clear, but also in contexts where there are many grey areas caused by metonymy. Metonymy is a figure of speech used in rhetoric in which a thing or concept is not called by its own name, but by the name of something intimately associated with that thing or concept. Metonyms can be either real or fictional concepts representing other concepts real or fictional, but they must serve as an effective and widely understood second name for what they represent. For example, (i)~\emph{Person vs. Artefact}: ``The Ham Sandwich (a person) wants his bill. vs ``Bring me a ham sandwich."; (ii)~\emph{Organization vs. Location}: ``England won the World Cup" vs. ``The World Cup took place in England"; (iii)~\emph{Company vs. Artefact}: ``shares in MTV" vs. ``watching MTV"; and (iv)~\emph{Location vs. Organization}: ``she met him at Heathrow" vs. ``the Heathrow authorities".

To address these challenges, the Message Understanding Conferences (MUC) were initiated and financed by DARPA (Defense Advanced Research Projects Agency) to encourage the development of new and better methods of information extraction. The tasks grew from producing a database of events found in newswire articles from one source to production of multiple databases of increasingly complex information extracted from multiple sources of news in multiple languages. The databases now include named entities, multilingual named entities, attributes of those entities, facts about relationships between entities, and events in which the entities participated. MUC essentially adopted simplistic approach of disregarding metonymous uses of words, e.g. `England' was always identified as a location. However, this is not always useful for practical applications of NER, such as in the domain of sports.

MUC defined basic problems in NER as follows: (i)~Variation of named entities: for example John Smith, Mr Smith, and John may refer to the same entity; (ii)~Ambiguity of named entities types: for example John Smith (company vs. person), May (person vs. month), Washington (person vs. location), and 1945 (date vs. time); (iii)~Ambiguity with common words: for example `may'; and (iv)~Issues of style, structure, domain, genre etc. as well as punctuation, spelling, spacing, and formatting. To address these challenges, the state of the art approaches to entity extraction proposed four primary steps~\cite{chen1998named,NER.evaluation.2009,nadeau2007survey,benjelloun2009swoosh}: Format Analysis, Tokeniser, Gazetteer, Grammar. Figure~\ref{NE_Process} illustrates a simplified process for the NER task. Following is brief description of these steps:

\begin{figure} [t]
\centering
  \includegraphics[scale=1.1]{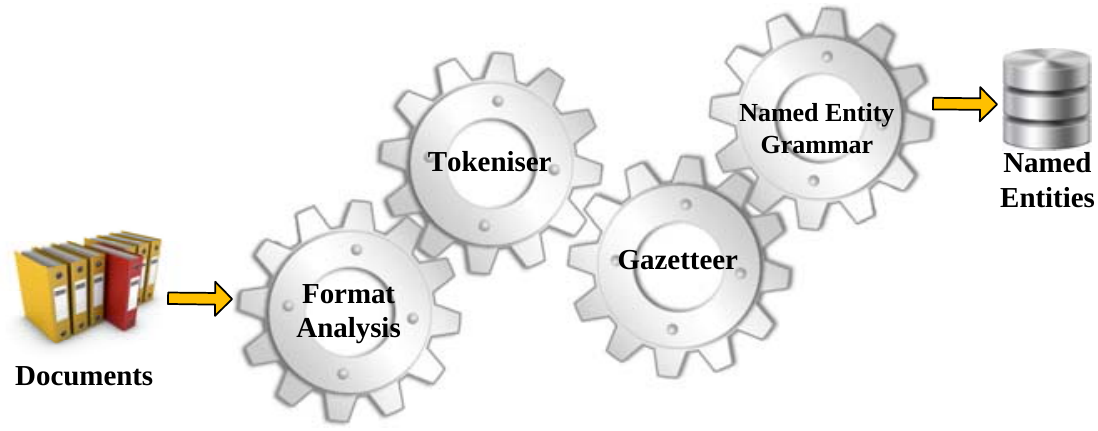}\\
  \caption{A simplified process for NER tasks.}\label{NE_Process}
\end{figure}


\paragraph{Format Analysis.} Many document formats contain formatting information in addition to textual content. For example, HTML documents contain HTML tags specifying formatting information such as new line starts, bold emphasis, and font size or style. The first step, format analysis, is the identification and handling of the formatting content embedded within documents that controls the way the document is rendered on a computer screen or interpreted by a software program. Format analysis is also referred to as structure analysis, format parsing, tag stripping, format stripping, text normalization, text cleaning, and text preparation.


\paragraph{Tokeniser.} Tokenization is the process of breaking a stream of text up into words, phrases, symbols, or other meaningful elements called tokens.This module is responsible for segmenting text into tokens, e.g., words, numbers, and punctuation. The list of tokens becomes input for further processing such as parsing or text mining.


\paragraph{Gazetteer.} This module is responsible for categorizing the type and scope of the information presented. In particular, a gazetteer is a geographical dictionary or directory, an important reference for information about places and place names. It typically contains information concerning the geographical makeup of a country, region, or continent as well as the social statistics and physical features, such as mountains, waterways, or roads. As an output, this module will generate set of named entities (e.g., towns, names, and countries) and key words (e.g., company designators and titles).


\paragraph{Grammar.} This module is responsible for hand-coded rules for named entity recognition. NER systems are able to use linguistic grammar-based techniques as well as statistical models. Hand-crafted grammar-based systems typically obtain better precision, but at the cost of lower recall and months of work by experienced computational linguists. Statistical NER systems typically require a large amount of manually annotated training~data.


\section{Entity Classification: State-of-the-Art}
\label{chap4}

The classification step is responsible for determining whether pairs of entities are co-referent or not.
To achieve this, extracted named entities in the entity identification phase should be compared by applying various features to pair of entities.
Such features can be divided into various classes such as string match~\cite{lee2011stanford,singh2011large,chen2012combining,wick2009entity}, lexical~\cite{chen1998named,bengtson2008understanding}, syntactic~\cite{skut1998chunk,tsuruoka2005developing}, pattern-based~\cite{daume2005large}, count-based~\cite{daume2005large,marquez2012coreference,potau2010coreference}, semantic~\cite{kambhatla2004combining,daume2005large}, knowledge-based~\cite{daume2005large,bryl2010using,nastase2010wikinet}, class-based~\cite{ravichandran2005randomized,pantel2009web,fastCoreference1,elsayed2008pairwise,sarmento2009approach}, list-/inference-/history-based~\cite{daume2005large}, and relationship-based~\cite{giles2008large,kambhatla2004combining} features. Table~\ref{tblFeature} illustrates the various classes of features, their description, and the state-of-the-art approaches.

\begin{table}
\begin{adjustwidth}{-0.9cm}{}
 \caption{Various classes of features.}
 \centering
  \begin{tabular}{cc}
   \includegraphics[scale=0.9]{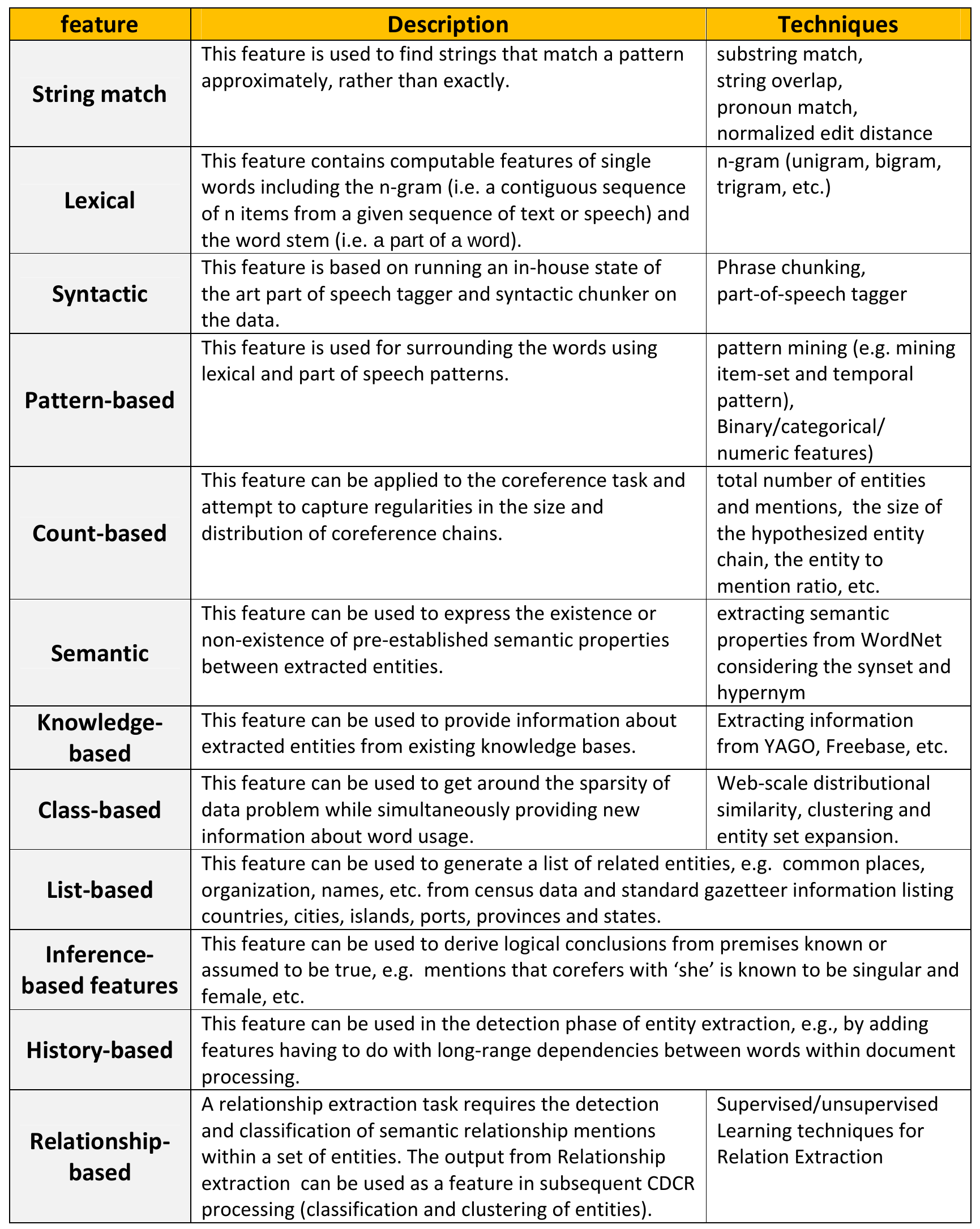}\\
  \end{tabular}
 \label{tblFeature}
\end{adjustwidth}
\end{table}

Recently, linked data~\cite{bizer2009linked} has become a prominent source of information about entities. Linked data describes a method of publishing structured data so that it can be interlinked and become more useful, and provides a publishing paradigm in which not only documents, but also data, can be a first class citizen of the Web. Projects such as DBpedia~\cite{DBpedia}, freebase~\cite{Freebase}, WikiTaxonomy~\cite{WikiTaxonomy}, and YAGO~\cite{Yago} have constructed huge knowledge bases (KBs) of entities, their semantic classes, and relationships among entities~\cite{BigDataMethods}. These systems can be used to enrich the entities with additional features and consequently to improve the effectiveness of the results. As an example, YAGO contains information harvested from Wikipedia, and linked to WordNet thesaurus~\cite{WordNet} as a semantic backbone, and having more than two million entities (e.g., people, organizations, and cities) and 20 million facts about these entities.

\subsection{Similarity Functions and Their Characteristics}

Approximate data matching usually relies on the use of a similarity function, where a similarity function $f(v_1, v_2) \mapsto s$ can be used to assign a score $s$ to a pair of data values $v_1$ and $v_2$. These values are considered to be representing the same real world object if $s$ is greater then a given threshold $t$.
In the classification step, similarity functions play a critical role in dealing with data differences caused by various reasons, such as misspellings, typographical errors, incomplete information, lack of standard formats, and so on. For example, personal name mentions may refer to a same person, but can have multiple conventions (e.g., \emph{Barack Obama} versus \emph{B. Obama}).

In the last four decades, a large number of similarity functions have been proposed in different research communities, such as statistics, artificial intelligence, databases, and information retrieval. They have been developed for specific data types (e.g., string, numeric, or image) or usage purposes (e.g., typographical error checking or phonetic similarity detection). For example, they are used for comparing strings (e.g., edit distance and its variations, Jaccard similarity, and tf/idf based cosine functions), for  numeric values (e.g., Hamming distance and relative distance), for  phonetic encoding (e.g., Soundex and NYSIIS), for  images (e.g., Earth Mover Distance), and so on. The functions can be categorized as follows.

\subsubsection{Similarity Functions for String Data}
\label{sec:stringTypeFunction}

For \textit{string} data types, in addition to exact string comparison, approximate string comparison functions~\cite{Hall1980} can be used for computing the similarity between two strings. They can be roughly categorized into three groups: \textit{character-based}, \textit{token-based} and \textit{phonetic} functions.

\paragraph{\bf Character-based Functions.}
These functions (e.g., edit distance, Jaro, or Jaro-Winkler) consider characters and their positions within strings to estimate the similarity~\cite{WangLF11}. Following we describe set of character-based functions.

\begin{description}
	\item \emph{Edit distance}: The edit distance between two strings is measured, based on the smallest number of edit operations (insertions, deletions, and substitutions) required to transform one string to the other. Each of the edit operations has a cost value (e.g., 1). For example, the edit distance between ``window'' and ``widow'' is 1 since deleting the letter ``n'' in ``window'' will convert the first string into the second. The edit distance function~\cite{Needleman} is expensive or less accurate for measuring the similarity between long strings (e.g., document or message contents). It is likely to be suitable for comparing short strings (e.g., document titles) capturing typographical errors or abbreviations~\cite{ElmagarmidIV07}.\\[-10pt]
	\item \emph{Jaro or Jaro-Winkler}: The Jaro function computes the string similarity by considering the number of common characters and transposed characters. Common characters are ones that emerge in both strings within half the length of the shorter string~\cite{tailor}. Transposed characters are ones that are non-matching when comparing common characters of the same position in both strings. The Jaro-Winkler function improves the Jaro function by using the length of the longest common prefix between two strings. These functions are likely to work well for comparing short strings (e.g., personal names).\\[-10pt]
	\item \emph{Q-grams}: Given a string, q-grams are substrings in which each consists of q characters of the string~\cite{Kukich1992}. For example, q-grams (q= 2) of ``susan'' are: `su', `us', `sa', and `an'. The similarity function computes the number of common q-grams in two strings and divides the number by either the minimum, average, or maximum number of q-grams of the strings~\cite{Christen06}. If strings have multiple words and tokens of the strings are ordered differently in another strings, the similarity function might be more effective than the other character-based functions, such as edit distance or jaro function~\cite{Christen06}.\\[-10pt]

\end{description}

\paragraph{\bf Token-based Functions.}
These functions might be appropriate in situations where the string mismatches come from rearrangement of tokens (e.g., ``James Smith'' versus ``Smith James'') or the length of strings is long, such as the content of a document or a message~\cite{KopckeTR10}. The following are some token-based functions:

\begin{description}
	\item \emph{Jaccard}: Jaccard function tokenizes two strings \texttt{s} and \texttt{t} into tokensets \texttt{S} and \texttt{T}, and quantifies the similarity based on the fraction of common tokens in the sets: $\frac{(S \cap T)}{(S \cup T)}$. For example, the jaccard similarity between ``school of computer science'' and ``computer science school'' is $\frac{3}{4}$. This function works well for the cases where word order of strings is unimportant.
	\item \emph{TF/IDF}: This function computes the closeness by converting two strings into unit vectors and measuring the angle between the vectors. In some situations, word frequency is important as in information retrieval applications that give more weight to rare words than on frequent words. In such cases, this function is likely to work better than the functions (e.g., Jaccard similarity) that are insensitive to the word frequency.
	
\end{description}

\paragraph{\bf Phonetic Similarity Functions.}
These functions describe how two strings are phonetically similar to each other in order to compute the string similarity. Some examples are as follows:

\begin{description}
	\item \emph{Soundex}~\cite{HolmesM02}, one of the best known phonetic functions, converts a string into a code according to an encoding table. A soundex code is comprised of one character and three numbers. The Soundex code is generated as follows: (i)~Keep the first letter of a string and ignore all other occurrences of vowels (a, e, i, o, u) and h, w, y; (ii)~Replace consonants with numbers according to Table~\ref{tab:encodingTable}; (iii)~Code two consecutive letters as a single number; and (iv)~Pad with 0 if there are less than three numbers.
For example, using the soundex encoding table, both ``daniel'' and ``damiel'' return the same soundex code ``d540''.
 \item \emph{Phonex/Phonix}~\cite{Randell93} is an alternative function to Soundex, which was designed to improve the encoding quality by preprocessing names based on their pronunciations. Phonix~\cite{Gadd1990}, an extension of Phonex, uses more than one hundred rules on groups of characters~\cite{Christen06}. The rules are applied to only some parts of names, e.g., the beginning, middle or end of names.
 \item \emph{Double Metaphone}~\cite{Philips} performs better for string matching in non-English languages, like European and Asian, rather than the soundex function that is suitable for English. Thus it uses more complex rules that consider letter positions as well as previous and following letters in a string, compared with the soundex function.
\end{description}

\begin{table}[t]
\centering
\begin{tabular}{|c|c|c|c|} \hline
\texttt{Consonants}& \texttt{Number} & \texttt{Consonants} & \texttt{Number} \\ \hline
b, f, p, v& 1  &  l & 4 \\ \hline
c, g, j, k, q, s, x, z& 2  &  m, n & 5 \\ \hline
d, t& 3  &  r & 6 \\ \hline
\end{tabular}
\vspace{2mm}
\caption{Soundex encoding table.}
\label{tab:encodingTable}
\end{table}

\subsubsection{Similarity Functions for Numeric Data}

For \textit{numeric} attributes, one can treat numbers as strings and then compare them using the similarity functions for string data described above or choose different functions for comparing numeric values as follows.

\begin{description}
	\item \emph{Relative Distance}: The relative distance is used for comparing numeric attributes $x$ and $y$ (e.g., price, weight, size): $R(x,y)= \frac{|x - y|}{max\{x,y\}}$. 
  \item \emph{Hamming Distance}:  The Hamming distance is the number of substitutions required to transform one number to the other. Unlike other functions (e.g., relative distance), it can be used only for comparing two numbers of equal length. For example, the Hamming distance between ``2121'' and ``2021'' is 1 as there is one substitution ($1\rightarrow 2$). The Hamming distance is used mainly for numerical fixed values, such as postcode and SSN~\cite{tailor}.
\end{description}

\subsubsection{Similarity Functions for Date or Time Data}

Date and time values must be converted to a common format in order to be compared with each other.
For example, possible formats for date type (considering day as `dd', month as `mm', and year as `yyyy'/`yy') include: `ddmmyyyy', `mmddyyyy', `ddmmyy', `mmddyy', and so on. For time type, times could be given as strings of the form `hhmm' or `mmhh' in 24 hours format. In the process during which date or time values are converted to a uniform representation, separator characters like `-', `/', `:' are removed from the values. To determine the similarity between these converted values, we could use numeric similarity functions (e.g., absolute difference) by considering them as numeric values or character-based similarity functions (e.g., edit distance) by considering them as string values.

\subsubsection{Similarity Functions for Categorical Data}

For \emph{categorical} features (whose values come from a finite domain), the similarity can be computed in a similar way to binary data types. For example, the score `1' is assigned for a match and the score `0' for a non-match. Alternatively, in~\cite{Anderberg1973}, the authors presented an approach that measures the similarity between two categorical values, based on user inputs. For example, Table~\ref{tab:categoricalData} shows the user-defined similarity scores between any two insurance types. This method can give more detailed scores between categorical data, rather than giving only two scores `0' or `1', although some effort is needed to provide user-defined similarity scores in advance.

\begin{table}
\centering
\begin{tabular}{|c|c|c|c|c|} \hline
Insurance Type& Car & Motorbike & Home & Travel  \\ \hline\hline
Car       &  1   &          &      &    \\ \hline
Motorbike &  0.7 & 1        &      &    \\ \hline
Home      &  0   & 0        & 1    &    \\ \hline
Travel    &  0   & 0        & 0.3  & 1   \\ \hline
\end{tabular}
\flushleft
\caption{Similarity scores between two insurance types.}
\label{tab:categoricalData}
\end{table}

Even though there is no similarity between any two feature values, further comparisons can be made because of \textit{semantic} relationships between them. For example, consider two feature strings ``Richard'' and ``Dick'' of person entities. Although normal string comparison functions may fail to see the similarity, the strings still can be considered as similar to each other, if we keep the information that the latter is an alias for the former.

\subsection{Clustering}


Clustering is the task of grouping a set of objects in such a way that objects in the same group (called cluster) are more similar (in some sense or another) to each other than to those outside the cluster. In information extraction, identifying the equivalence classes of entity mentions is the main focus: it is important that an entity  and all its mentions  are placed in the same equivalence class. In this context, the goal of coreference resolution will be to identify and connect all textual entity mentions that refer to the same entity.

To achieve this goal, it is important to identify all references within the same document (i.e. within document coreference resolution). An intra-document coreference system can be used to identify each reference and to decide if these references refer to a single individual or multiple entities. Some techniques (e.g. ~\cite{rao2010streaming,green2012entity}) create a coreference chain for each unique entity within a document and then group related coreference chains in similar clusters. Then, they use a streaming clustering approach with common coreference similarity computations to achieve high performance on large datasets. The proposed method is designed to support both entity disambiguation and name variation operating in a streaming setting, in which documents are processed one at a time and only once.

The state-of-the-art in clustering has been discussed in previous publications~\cite{GooiA04,sekine2009named,nadeau2007survey}. Many of these approaches rely on mention (string) matching, syntactic features, and linguistic resources like English WordNet~\cite{stark1998wordnet}. The classic works on clustering~\cite{bagga1998algorithms,GooiA04} adapted the Vector Space Model (VSM\footnote{Vector space model or term vector model is an algebraic model for representing text documents (and any objects, in general) as vectors of identifiers}) or deployed different information retrieval techniques for entity disambiguation and clustering. Such works showed that clustering documents by their domain specific attributes such as domain genre will affect the effectiveness of cross-document coreferencing.

Some extensions to VSM for for cross-document coreference clustering have been proposed in~\cite{mann2003unsupervised,chen2007towards}. Furthermore, supervised approaches~\cite{black1998facile}, Semi-supervised approaches~\cite{ando2005high}, and and unsupervised approaches~\cite{elsner2009structured} have used clustering to group together different nominal referring to the same entity.
In particular, approaches to cross document coreference resolution have first constructed a vector space representation derived from local (or global) contexts of entity mentions in documents and then performed some form of clustering on these vectors. Most of these approaches focused on disambiguating personal names.

Another line of related work, e.g.~\cite{fleischman2004multi,MayfieldADEE09} added a discriminative pairwise mention classifier to a VSM-like model. For example, Mayfield et al.~\cite{MayfieldADEE09}, clustered the resulting entity pairs by eliminating any pair with an SVM output weight of less than 0.95, then they treated each of the connected components in the resulting graph as a single entity. Ah-Pine et al.~\cite{ah2009clique} proposed a clique-based clustering method based upon a distributional approach which allows to extract, analyze and discover highly relevant information for corpus specific NEs annotation. Another line of related work~\cite{green2011entity,aktolga2008cross} proposed techniques for clustering text mentions across documents and languages simultaneously. Such techniques may produce cross-lingual entity clusters. Some later work~\cite{ni2010enhancing,attardi2010tanl} relies on the use of extremely large corpora which allow very precise, but sparse features. For example, Ni et al.~\cite{ni2010enhancing} enhanced the open-domain classification and clustering of named entity using linked data approaches.

Dynamic clustering approach [9] follows the method in which set of points are observed from a potentially infinite set X, one at a time, in order to maintain a fixed number of clusters while minimizing the maximum cluster radius (i.e. the radius of the smallest ball containing all points of the cluster). This approach consists of two stages: update and merge. Update adds points to existing clusters or creates new clusters while merge combines clusters to prevent the clusters from exceeding a fixed limit. Comparing to the agglomerative clustering approach (which has the quadratic cost), the streaming clustering provides a potentially linear performance in the number of observations since each document need only be examined a single time.

\section{CDCR and Big Data}
\label{chap5}

The ability to harness the ever increasing amounts of data will enable us to understand what is happening in the world. In this context, big data enables the two main characteristics to come together: (i)~big data for massive amounts of detailed information; and (ii)~advanced analytics including artificial intelligence, natural language processing, data mining, statistics and so on. Generating huge metadata for imbuing the data with additional semantics will form part of the big data challenges in CDCR. For example, `Barack Obama' can be a student in the Harvard Law School in a period of time and can be the president of the United States in another time. More specifically, entities and their mentions may have massive amounts of detailed information which need to be analyzed over time.

Big data has raised various challenges in different tasks of information extraction, including in CDCR. In the entity identification phase, entity extraction subtasks such as format analysis, tokeniser, gazetteer, and grammar would have to be applied to huge number of documents. This is challenging as, in terms of scalability, entity extraction outputs more data than it takes. For example, as illustrated in Table~\ref{fig:dataset}, only 6600 documents provide more than two million entities. In contrast, the English Gigaword dataset contains more that nine million documents and will produce orders of magnitude more information.

Currently, the dominant methods for co-reference measure compatibility between pairs of mentions. These suffer from a number of drawbacks including difficulties scaling to large numbers of mentions and limited representational power~\cite{fastCoreference1}. For example, as illustrated in Table~\ref{fig:dataset}, for more than 30,000 extracted named entities, around 900 million pairs can be generated. In particular, in terms of scalability, pairwise entity comparison will become exponential across documents.

Recent research~\cite{fastCoreference1,wellner2004integrated,ng2010supervised,wick2009entity} have studied methods that measure compatibility between mention pairs (i.e., the dominant approach to coreference) and showed that these approaches suffer from a number of drawbacks including difficulties scaling to large numbers of mentions and limited representational power. For example, Wick et al.~\cite{fastCoreference1} proposed to replace the pairwise approaches with a more expressive and highly scalable alternatives, e.g., discriminative hierarchical models that recursively partitions entities into trees of latent sub-entities. Wellner et al.~\cite{wellner2004integrated} proposed an approach to integrated inference for entity extraction and coreference based on conditionally-trained undirected graphical models. 
Luo et al.~\cite{luo2004mention} proposed an approach for coreference resolution which uses the Bell tree to represent the search space and casts the coreference resolution problem as finding the best path from the root of the Bell tree to the leaf nodes.
Wick et al.~\cite{wick2009entity} proposed a discriminatively-trained model that jointly performs coreference resolution and canonicalization, enabling features over hypothesized entities.

Finally, in the classification step, various similarity metrics should be calculated for all generated paired entities, and then the huge number of coreferent entities should be clustered and placed in the same equivalence class. To address these challenges, and to effectively classify and cluster these gigantic number of entities and pairs, parallel and distributed architectures have become popular. MapReduce~\cite{MapReduce} is a distributed computing framework introduced by Google with the goal of simplifying the process of distributed data analysis. \

The MapReduce programming model consists of two functions called Map and Reduce. Data are distributed as key-value pairs on which the Map function computes a different set of intermediate key and value pairs. An intermediate Shuffle phase groups the values around common intermediate keys. The Reduce function then performs computation on the lists of values with the same key. The resulting set of key-value pairs from the reducers is the final output. Apache Hadoop~\cite{Hadoop} is the most popular, open source implementation of MapReduce that provides a distributed file system (i.e., HDFS\footnote{http://hadoop.apache.org/}) and also, a high level language for data analysis, i.e., Pig\footnote{http://pig.apache.org/}. Hadoop~\cite{Hadoop} can be used to build scalable algorithms for pattern analysis and data mining. This has been demonstrated by recent research~\cite{elsayed2008pairwise,pantel2009web,sarmento2009approach,singh2011large,kolb2012dedoop} that have used MapReduce~\cite{MapReduce} for processing huge amounts of documents in a massively parallel way.


%
%
Elsayed et al.~\cite{elsayed2008pairwise} proposed a MapReduce algorithm for computing pairwise document similarity in large document collections. The authors focused on a large class of document similarity metrics that can be expressed as an inner product of term weights. They proposed a two step solution to the pairwise document similarity problem: (i)~Indexing, where a standard inverted index algorithm~\cite{frakes1992information} in which each term is associated with a list of document identifiers for documents that contain it and the associated term weight; and (ii)~Pairwise Similarity, where the MapReduce mapper generates key tuples corresponding to pairs of document IDs in the postings in which the key tuples will be associated with the product of the corresponding term weights.

A scalable MapReduce-based implementation based on distributional similarity have been proposed in~\cite{pantel2009web}, where the approach followed a generalized sparse-matrix multiplication algorithm~\cite{sarawagi2004efficient}. The MapReduce plan uses the Map step to start $M*N$ Map tasks in parallel, each caching $1/M_{th}$ part of $A$ as an inverted index and streaming $1/N_{th}$ part of $B$ through it. In this approach, there is a need to process each part of $A$ for $N$ times, and each part of $B$ is processed $M$ times.

A multi-pass graph-based clustering approach to large scale named-entity disambiguation have been proposed in~\cite{sarmento2009approach}. The proposed MapReduce-based algorithm is capable of dealing with an arbitrarily high number of entities types is able to handle unbalanced data distributions while producing correct clusters both from dominant and non-dominant entities. Algorithms will be applied to constructed clusters for assigning small clusters to big clusters, merging small clusters, and merging big and medium clusters. According to these related works, MapReduce Algorithm design could lead to data skew and the curse of the last reducer and consequently careful investigation is needed while mapping an algorithm into the MapReduce plan.

A distributed inference that uses parallelism to enable large scale processing have been proposed in~\cite{singh2011large}. The approach uses a hierarchical model of coreference that represents uncertainty over multiple granularities of entities. The approach facilitates more effective approximate inference for large collections of documents. They divided the mentions and entities among multiple machines, and propose moves of mentions between entities assigned to the same machine. This ensures all mentions of an entity are assigned to the same machine.
Kolb et al.~\cite{kolb2012dedoop} proposed a tool called Dedoop (Deduplication with Hadoop) for MapReduce-based entity resolution of large datasets. Dedoop automatically transforms the entity resolution workflow definition into an executable MapReduce workflow. Moreover, it provides several load balancing strategies in combination with its blocking techniques to achieve balanced workloads across all employed nodes of the cluster.

\section{CDCR Tools and Techniques Evaluation}
\label{sec6}

\subsection{Evaluation Dimensions}
\label{chapEval}

Cross-Document Coreference Resolution (CDCR) is the task of identifying entity mentions (e.g., persons, organizations or locations) across multiple documents that refer to the same underlying entity. An important problem in this task is how to evaluate a system's performance. 

There are two requirements that should be lie at the heart of CDCR task: (i)~effectiveness, which concerns with achieving a high quality coreference result. For the evaluation of accuracy, well-known measures such as \emph{precision} (the fraction of retrieved instances that are relevant) and \emph{recall} (the fraction of relevant instances that are retrieved)~\cite{salton1986introduction} can be used; and (ii)~efficiency, that concerns with performing the coreference resolution as fast as possible for large datasets. 

In this context, a good performance metric should have the following two properties~\cite{luo2005coreference}:

\begin{description}
  \item \emph{discriminativity}: which is the ability to differentiate a good system from a bad one. For example, precision and recall have been proposed to measure the effectiveness of information retrieval and extraction tasks, where high recall means that an algorithm returned most of the relevant results and  high precision means that an algorithm returned substantially more relevant results than irrelevant;
  \item \emph{interpretability}: which emphasis that a good metric should be easy to interpret. In particular, there should be an intuitive sense of how good a system is when a metric suggests that a certain percentage of coreference results are correct. For example, when a metric reports 95\% or above correct for a system, we would expect that the vast majority of mentions are in right entities or coreference chains;
\end{description}

For the evaluation of accuracy, well-known measures such as precision (the fraction of retrieved instances that are relevant) and recall (the fraction of relevant instances that are retrieved)~\cite{salton1986introduction} can be used. As the complementary to precision/recall, some approaches such as link-based F-measure~\cite{vilain1995model}, count the number of common links between the truth (or reference) and the response. In these approaches, the link precision is the number of common links divided by the number of links in the system output, and the link recall is the number of common links divided by the number of links in the reference. The main shortcoming of these approaches is that they fail to distinguish system outputs with different qualities: they may result in higher F-measures for worse systems.

Some other value-based metric such as ACE-value~\cite{nist2003ace} count the number of false-alarm (the number of miss) and the number of mistaken entities. In this context, they associate each error with a cost factor that depends on things such as entity type (e.g., location and person) as well as mention level (e.g., name, nominal, and pronoun). 
The main shortcoming of these approaches is that they are hard to interpret. For example a system with 90\% ACE-value does not mean that 90\% of system entities or mentions are correct: the cost of the system, relative to the one outputting zero entities is 10\%. To address this shortcoming, approaches such as Constrained Entity-Aligned F-Measure (CEAF)~\cite{luo2005coreference} have been proposed to measure the quality of a coreference system where an intuitively better system would get a higher score than a worse system, and is easy to interpret.

B-cubed metric~\cite{bagga1998algorithms}, a widely used approach, proposed to address the aforementioned shortcomings by first computing a precision and recall for each individual mention and then taking the weighted sum of these individual precisions and recalls as the final metric. The key contributions of this approach include: promotion of a set-theoretic evaluation measure, B-CUBED, and the use of TF/IDF\footnote{tf/idf, term frequency/inverse document frequency, is a numerical statistic which reflects how important a word is to a document in a collection or corpus.} weighted vectors and `cosine similarity'\footnote{Cosine similarity is a measure of similarity which can be used to compare entities/documents in text mining. In addition, it is used to measure cohesion within clusters in the field of data mining.} in single-link greedy agglomerative clustering. In particular, B-Cubed looks at the presence/absence of entities relative to each of the other entities in the equivalence classes produced: the algorithm computes the precision and recall numbers for each entity in the document, which are then combined to produce final precision and recall numbers for the entire output.

{\subsection{Datastes}
\label{sec:datasets}}

Measuring the effectiveness of CDCR task on large corpuses is challenging and needs large datasets providing sufficient level of ambiguity (the ability to express more than one interpretation) and sound ground-truth (the accuracy of the training set's classification for supervised learning techniques). Several manually/automatically labeled datasets have been constructed for training and evaluation of coreference resolution methods, however,
CDCR supervision will be challenging as it has a exponential hypothesis space in the number of mentions. Consequently, the manual annotation task will be time-consuming, expensive and will result in few number of ground-truths.

Few publications~\cite{bentivogli2008creating,day2008corpus,Bagga1998} introduced manually-labeled, small datasets containing high ambiguity, which make it very hard to evaluate the effectiveness of the CDCR techniques. Several automatic methods for creating CDCR datasets have been proposed to address this shortcoming. For example, recently, Google released the Wikilinks Corpus\footnote{http://googleresearch.blogspot.com.au/2013/03/learning-from-big-data-40-million.html}~\cite{singh12:wiki-links} which includes more than 40 million total disambiguated mentions over 3 million entities within around 10 million documents. Other examples of automatically labeled large datasets includes ~\cite{niu2004weakly,GooiA04,Sameerabs2010,spitkovsky2012cross}. Following we provide more details about TAC-KBP, John Smith, ACE, reACE, English Gigaword, and Google's Wikilinks datasets.
%

\begin{description}
\item[{\bf John Smith corpus~\cite{Bagga1998}.}] This dataset is one of the first efforts for creating corpora to train and evaluate cross-document co-reference resolution algorithms. The corpus is a highly ambiguous dataset which consisted of 197 articles from 1996 and 1997 editions of the
New York Times. The relatively of common name `John Smith' used to find documents that were about different individuals in the news.

\item [{\bf ACE (2008) corpus~\cite{extraction2008evaluation}.}] The most recent Automatic Content Extraction (ACE) evaluation took place in 2008, where the dataset includes approximately 10,000 documents from several genres (predominantly newswire). As the result of ACE participation (participants were expected to cluster person and organization entities across the entire collection), a selected set of about 400 documents were annotated and used to evaluate the system performance.

\item [{\bf reACE~\cite{hachey2012datasets}.}] The dataset was developed at the University of Edinburgh which consists of English broadcast news and newswire data originally annotated for the ACE (Automatic Content Extraction) program to which the Edinburgh Regularized ACE (reACE) mark-up has been applied. In order to provide a sufficient level of ambiguity and reasonable ground-truth, the dataset includes annotation for: (1)~a refactored version of the original data to a common XML document type; (2)~linguistic information from LT-TTT (a system for tokenizing text and adding markup) and MINIPAR (an English parser); and (3)~a normalized version of the original RE markup that complies with a shared notion of what constitutes a relation across domains. Similar to ACE and John Smith corpus, this dataset contains few annotated documents and cannot be used to evaluate the efficiency of big-data approaches in CDCR.

\item [{\bf English Gigaword}] is a comprehensive archive of newswire text data that has been acquired over several years by the LDC at the University of Pennsylvania. The fifth edition of this dataset includes seven distinct international sources of English newswire and contains more than 9 million documents. This large dataset is not annotated but can be used to assess the efficiency of the CDCR approaches.

\item [{\bf Google's Wikilinks Corpus~\cite{singh12:wiki-links}}] This dataset comprises of 40 million mentions over 3 million entities gathered using an automatic method based on finding hyperlinks to Wikipedia from a web crawl and using anchor text as mentions~\cite{singh12:wiki-links}. The Google search index has been used to discover the mentions that belong to the English language. The dataset provides the URLs of all the pages that contain labeled mentions, the actual mentions (i.e., the anchor text), the target Wikipedia link (entity label), and the byte offsets of the links on the page. Similar to Wikilinks, the \emph{TAC-KBP corpus}~\cite{mcnamee2009overview} links entity mentions to corresponding Wikipedia derived knowledge base nodes, focusing on ambiguous person, organization, and geo-political entities mentioned in newswire, and required systems to cope with name variation and name disambiguation. The dataset contains over 1.2 million documents, primarily newswire.
\end{description}

\subsection{Tools for Entity Identification and their evaluation}

In this section, we assess a set of named entity extraction systems including: OpenNLP, Stanford-NLP, LingPipe, Supersense tagger, AFNER, and AlchemyAPI.
Table~\ref{tools} illustrates a set of Information Extraction tools and their applications.
Table~\ref{tools_step} depicted CDCR tasks and the tools that can be leveraged in each phase.
The assessment only consider the names of persons, locations and organizations. The motivation behind this assessment is to provide a complementary vision for the results of domain independent systems that permit the processing of texts as well as process texts in a common language: English has been the selected language for this assessment. Following is a brief description of the selected tools.

\paragraph{\bf Stanford-NLP}\footnote{http://nlp.stanford.edu/},
is an integrated suite of natural language processing tools for English in Java, including tokenization, part-of-speech tagging, named entity recognition, parsing, and coreference. Stanford NER provides a general implementation of linear chain Conditional Random Field (CRF) sequence models, coupled with well-engineered feature extractors for Named Entity Recognition. The model is dependent on the language and entity type it was trained for and offers a number of pre-trained name finder models that are trained on various freely available corpora.

\paragraph{\bf OpenNLP}\footnote{http://opennlp.apache.org/},
is a machine learning based toolkit for the processing of natural language text. It supports the most common NLP tasks, such as tokenization, sentence segmentation, part-of-speech tagging, named entity extraction, chunking, parsing, and coreference resolution. The OpenNLP Name Finder can detect named entities and numbers in text. The Name Finder needs a model to be able to detect entities. The model is dependent on the language and entity type it was trained for. The OpenNLP projects offers a number of pre-trained name finder models that are trained on various freely available corpora. The OpenNLP engine reads the text content and leverages the sentence detector and name finder tools bundled with statistical models trained to detect occurrences of named entities.

The OpenNLP tools are statistical NLP tools including a sentence boundary detector, a tokenizer, a POS tagger, a phrase chunker, a sentence parser, a name finder and a coreference resolver. The tools are based on maximum entropy models. The OpenNLP tools can be used as standalone (in which the output will be a single text format) or as plugins with other Java frameworks including UIMA (in which the output will be in XML metadata Interchange (XMI) format). It is possible to pipe output from one OpenNLP tool into the next, e.g., from the sentence detector into the tokenizer.

\begin{table}
\begin{adjustwidth}{-2cm}{}
 \caption{List of existing Information Extraction tools and their applications.}
 \centering
  \begin{tabular}{cc}
   \includegraphics[scale=0.9]{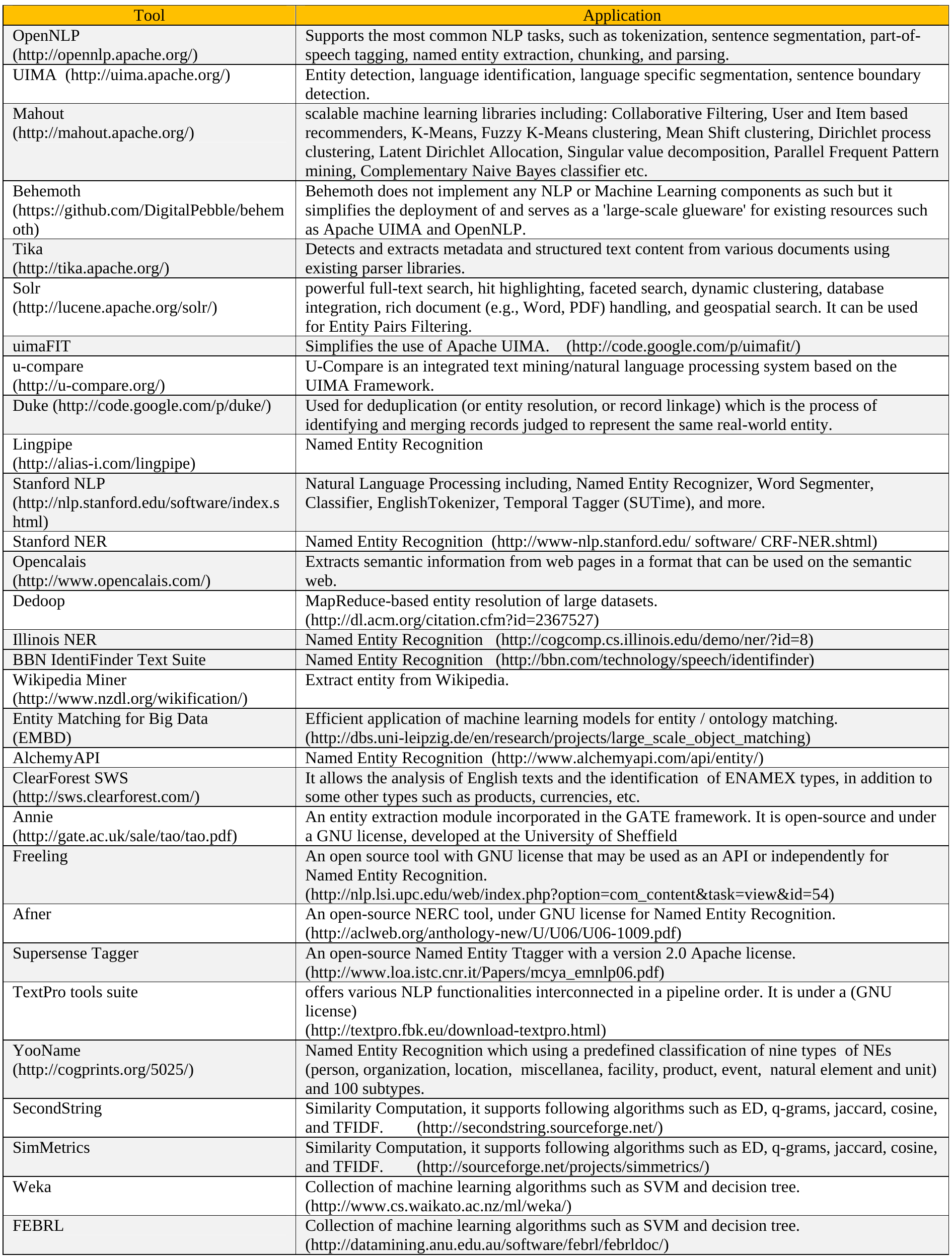}\\
  \end{tabular}
 \label{tools}
\end{adjustwidth}
\end{table}

\begin{landscape}
\begin{table}
\begin{adjustwidth}{-0.5cm}{}
 \caption{CDCR tasks and the tools which can be leveraged in each phase.}
 \centering
  \begin{tabular}{cc}
   \includegraphics[scale=0.9]{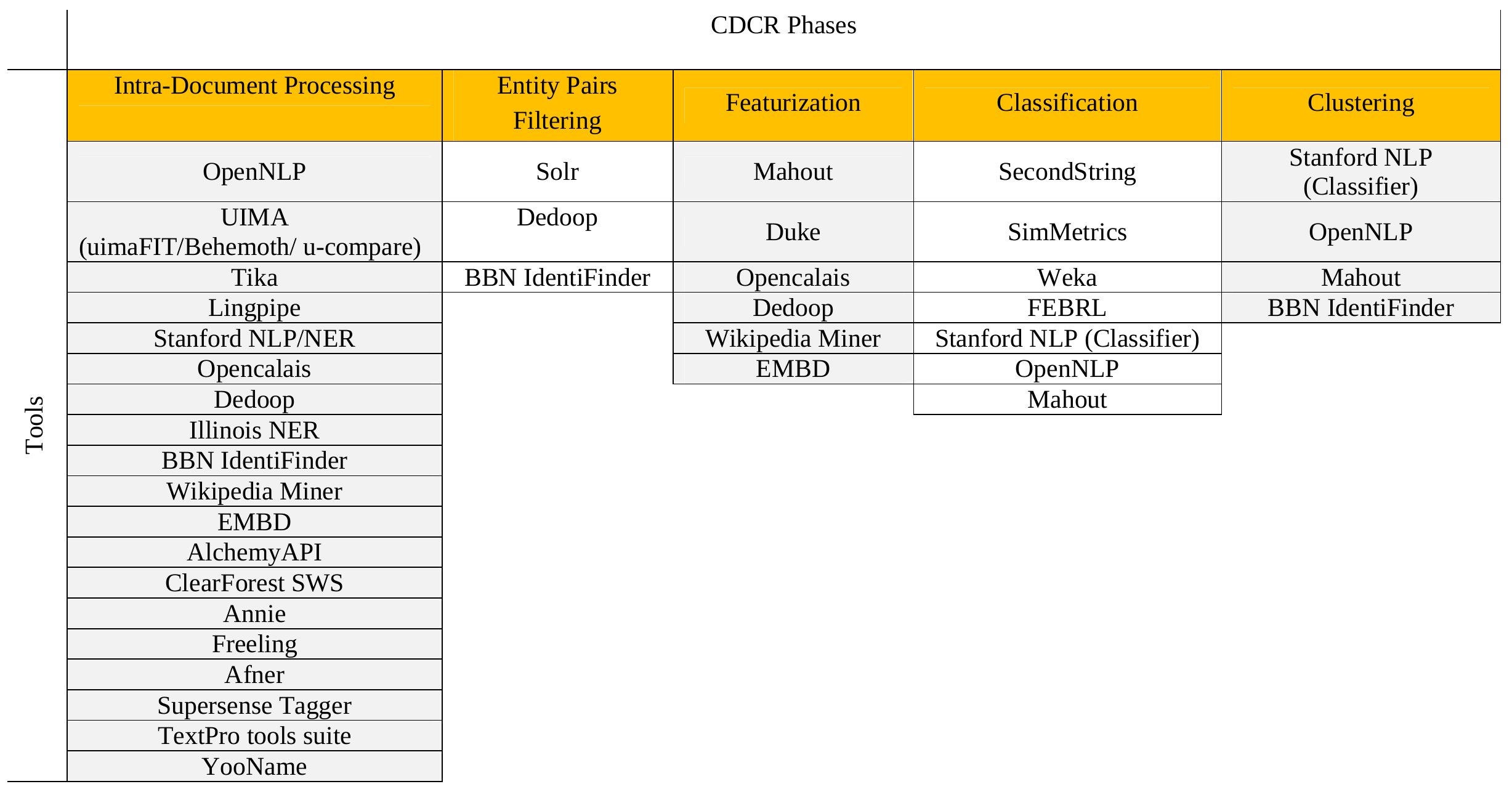}\\
  \end{tabular}
 \label{tools_step}
\end{adjustwidth}
\end{table}
\end{landscape}

The OpenNLP sentence detector is based on the approach proposed in~\cite{OpenNLPSentence}. One obvious drawback in the classification approach is that it cannot identify sentence boundaries where there is no marker. Next step is the statistical tagging. The statistical approach to tagging is to treat it as a multi-way classification problem. The OpenNLP POS (Part of speech) tagger is based on the approach proposed in~\cite{OpenNLP_POS}. After the OpenNLP tagger was developed, Toutanova and Manning~\cite{StanfordNLP_POS} proposed approaches for improving the accuracy of maximum entropy taggers. The Stanford-NLP POS (Part of speech) is based on this latter~work.

Chunking (also known as partial parsing) creates very shallow trees representing simple, flat phrase structure (mainly noun phrases). The basic approach in chunking is to exploit the work already done by the POS tagger in order to identify simple phrases by recognizing sequences of POS tags.
The OpenNLP tools include a chunker, which uses a maximum entropy model to recognize patterns in the POS tags made by the OpenNLP tagger. Stanford NLP does not provide chunker.
The Stanford parser is actually a set of alternative parsing algorithm and statistical models. It was developed in order to compare and evaluate different techniques.

\paragraph{\bf LingPipe}\footnote{http://alias-i.com/lingpipe/},
is a toolkit for processing text using computational linguistics. LingPipe is used to detect named entities in news, classify Twitter search results into categories, and suggest correct spellings of queries.
It includes multi-lingual, multi-domain, and multi-genre models as well as training with new data for new tasks. Moreover, it includes online training (learn-a-little, tag-a-little) and character encoding-sensitive I/O.
It offers a user interface and various demos through which it is possible to test texts. We used the latest release of LingPipe,  LingPipe~4.1.0, in the assessment.

\paragraph{\bf Supersense Tagger}\footnote{https://sites.google.com/site/massiciara/},
is designed for the semantic tagging of nouns and verbs based on WordNet categories which includes set of named entities such as persons, organizations, locations, temporal expressions and quantities. It is based on automatic learning, offering three different models for application: CONLL, WSJ and WNSS. The Supersense-CONLL have been used in our evaluation.

\paragraph{\bf AFNER}\footnote{http://afner.sourceforge.net/afner.html},
is a package for named entity recognition. AFNER uses regular expressions to find simple case named entities such as simple dates, times, speeds, etc. Moreover, it supports finding the parts of text matching listed named entities. The regular expression and list matches are then used in a `\emph{maximum entropy}'\footnote{Maximum entropy is a probability distribution estimation technique widely used for a variety of natural language tasks, such as language modeling, part-of-speech tagging, and text segmentation. Maximum entropy can be used for text classification by estimating the conditional distribution of the class variable given the document.} based classifier. Features relating to individual tokens (including list and regular expression matches) as well as contextual features are used. It also allows the addition of lists and regular expressions, as well as the training of new models. It is by default
capable of recognizing persons' names, organizations, locations, miscellanies, monetary quantities, and dates in English texts.

\paragraph{\bf AlchemyAPI}\footnote{http://www.alchemyapi.com/},
utilizes natural language processing technology and machine learning algorithms to analyze content, extracting semantic meta-data: information about people, places, companies, topics, facts and relationships, authors, languages, and more. API endpoints are provided for performing content analysis on Internet-accessible web pages, posted HTML or text content. It supports multiple languages and offers comprehensive disambiguation capabilities solutions. Moreover, it can be used to identify positive, negative and neutral sentiment within HTML pages and text documents/contents as well as for extracting document-level sentiment, user-targeted sentiment, entity-level sentiment, and keyword-level sentiment.

\subsubsection{Analysis and Methodology}

\begin{table}
 \caption{Main characteristics of the datasets.}
 \centering
  \begin{tabular}{cc}
   \includegraphics[scale=0.62]{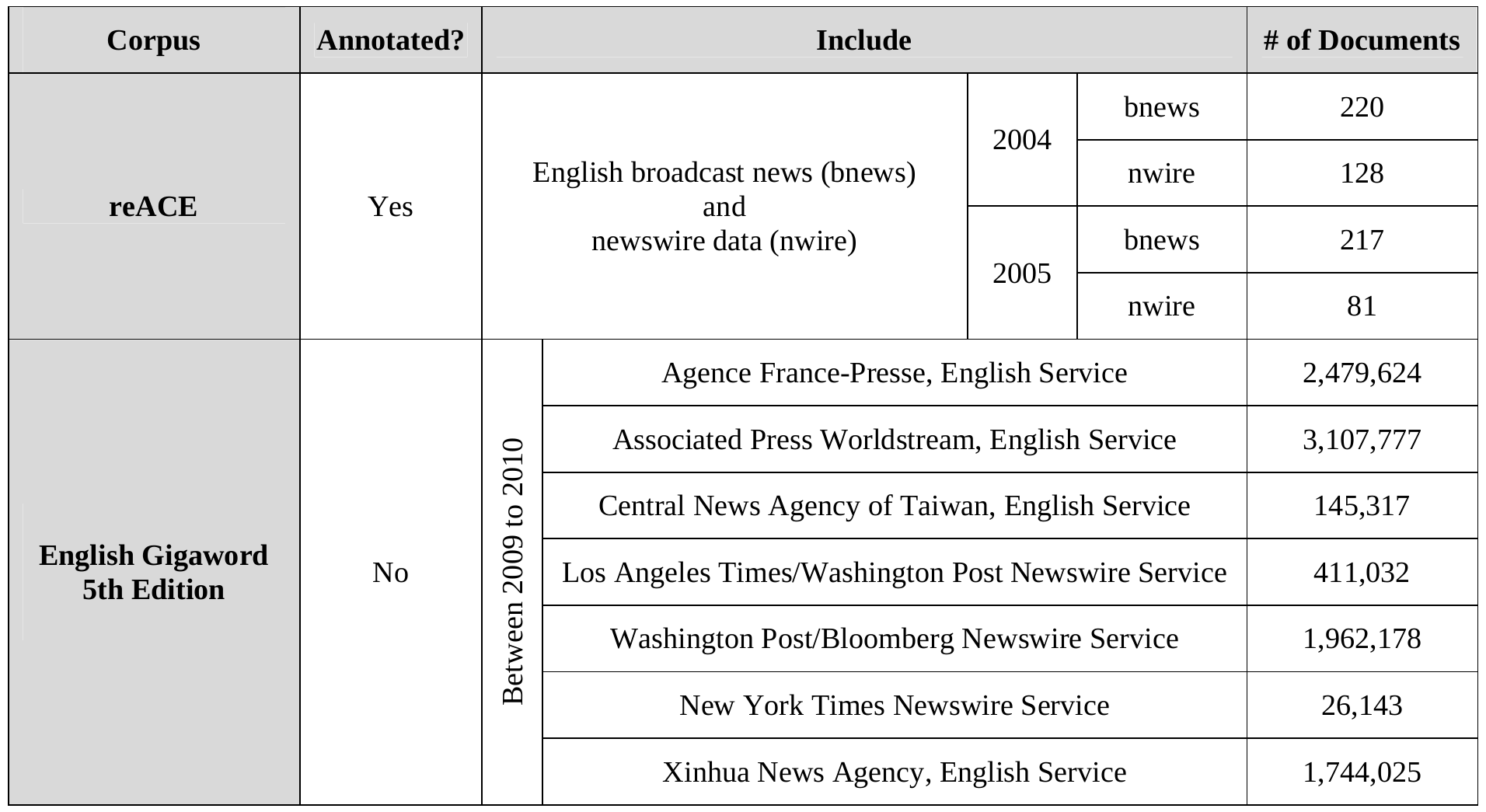}\\
  \end{tabular}
 \label{datasets}
\end{table}

In this assessment, we use reAce (to evaluate the effectiveness of the results) and English Gigaword (to evaluate the efficiency of the results) datasets. These datasets have been discussed in Section~\ref{sec:datasets}. Table~\ref{datasets} illustrates the main characteristics of these datasets.
The data analysis has been realized having focused on comparison of results obtained by the tools, for entities found in the test corpus: set of documents in the English Gigaword corpus has been used in order to evaluate the behavior of the tools. In particular, we used part of the Agence France-Presse, English Service (\texttt{afp\_eng}), as part of English Gigaword corpus, that has a total of 492 words, distributed in 5 documents and 21 paragraphs in which more than 60 occurrences of various types of entities have been accumulated. The assessment only consider the names of persons, locations and organizations. These entity types were distributed in various phrases in the corpus with different typography, where entities in a tool could neither totally coincide in number nor in semantic with their equivalent entities in other tools. Consequently, we adopted the corpus for every tool.

The data analysis has been realized having focused on the comparison of results obtained by the tools, for entities found in the test corpus. For the evaluation of accuracy, we use the well-known measures of precision and recall~\cite{salton1986introduction}. As discussed earlier, precision is the fraction of retrieved instances that are relevant, while recall is the fraction of relevant instances that are retrieved. In particular, precision measures the quality of the matching results, and is defined by the ratio of the correct entities to the total number of entities found:

\begin{center}
Precision = $\frac{number-of-currect-entities-found}{total-number-of-entities-extracted}$
\end{center}

Recall measures coverage of the matching results, and is defined by the ratio of the correct entities matched to the total number of all correct entities that should be found.

\begin{center}
Recall = $\frac{number-of-currect-entities-found}{total-number-of-correct-entities-that-should-be-ound}$
\end{center}

\begin{figure} [t]
\centering
\includegraphics[width=0.75\textwidth]{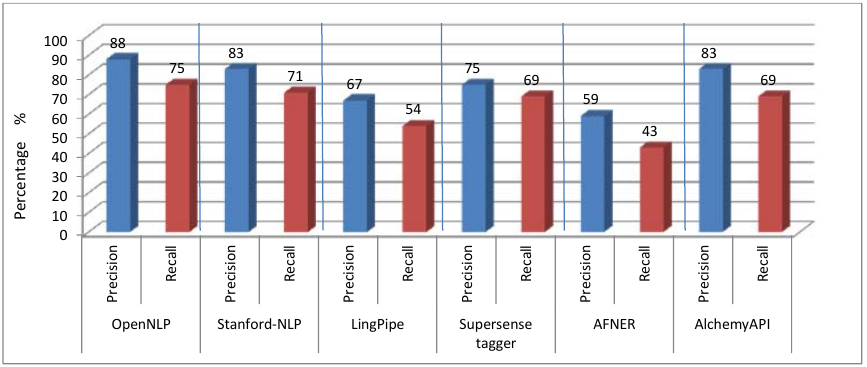}
\caption{Precision-Recall in entity identification.}
\label{fig:PR-identification}
\end{figure}

For an approach to be effective, it should achieve a high precision and high recall. However, in reality these two metrics tend to be inversely related~\cite{salton1986introduction}. The evaluation has been realized through distinct measures of precision and recall based on: (i)~ identification of the entities and false-positives\footnote{In statistics, a false positive is the incorrect rejection of a true null hypothesis, which may lead one to conclude that a thing exists when really it doesn't. For example, that a named entity is of a specific type, e.g. Person, when the entity is not of that~type.} in the identification; (ii)~classification of entities; and (iii)~classification by NE types that each tool recognizes.

\begin{figure} [t]
\centering
\includegraphics[width=0.75\textwidth]{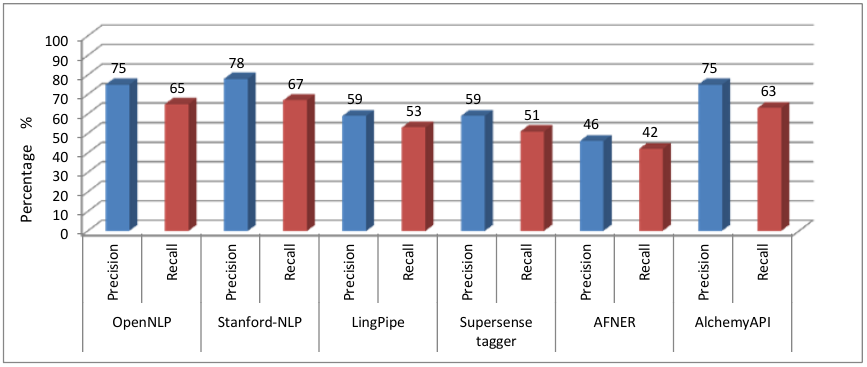}
\caption{Precision-Recall in entity classification.}
\label{fig:PR-classification}
\end{figure}

Figure~\ref{fig:PR-classification} illustrates the precision-recall for entity classification. Notice that, entity classification is the process of classifying entities based on their type (i.e., recognized by the tools) and is different from coreference classification (see Section~4).
Given that classification is a process that depends on the identification of entities, the f-measure in identification is always superior to that of the classification. In particular, F-measure is the harmonic mean of precision and recall:

\begin{center}
F-measure=$2.\frac{precision.recall}{precision+recall}$
\end{center}

\begin{figure} [t]
\centering
\includegraphics[width=0.75\textwidth]{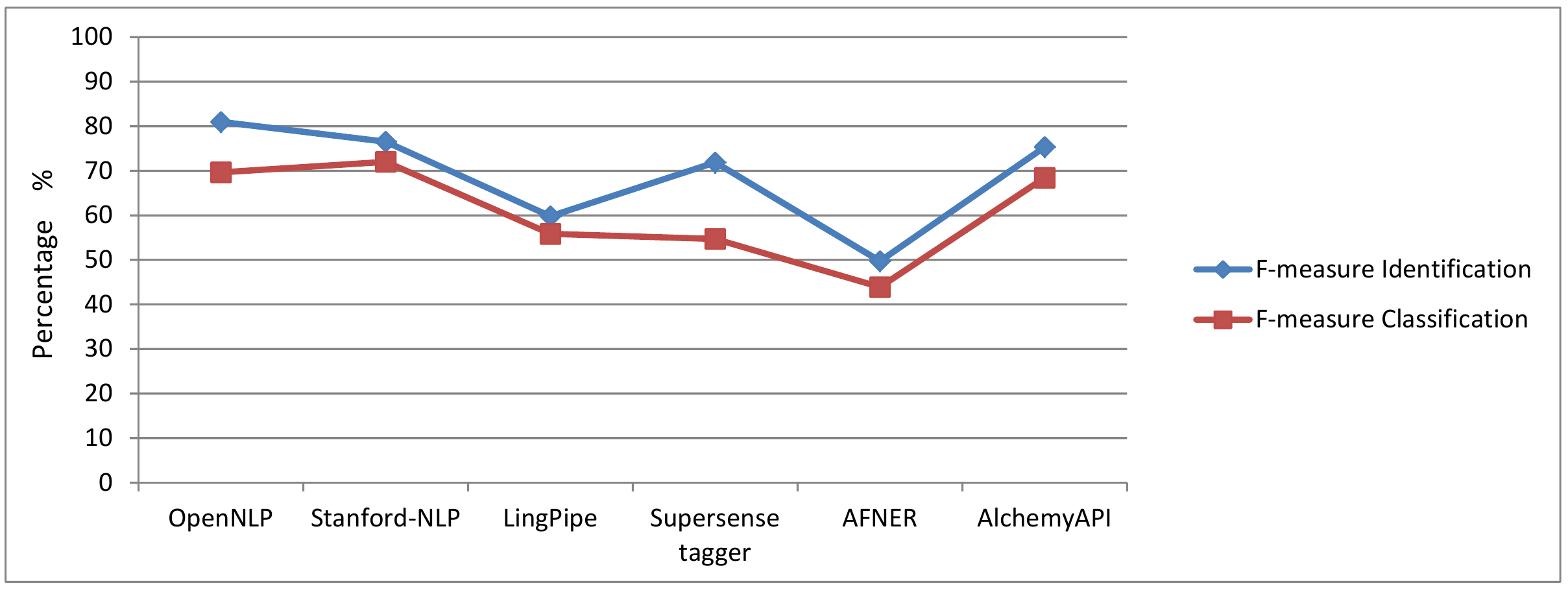}
\caption{F-measure in entity identification and classification.}
\label{fig:PR-fmeasure}
\end{figure}

Figure~\ref{fig:PR-fmeasure} illustrates the F-measure in entity identification and classification.
Comparing to the precision-recall for entity identification and classification, it is generally observed that the values are similar for the F-measure in entity identification and classification.


\begin{table}
 \caption{Results by entity type.}
 \centering
  \begin{tabular}{cc}
   \includegraphics[scale=0.5]{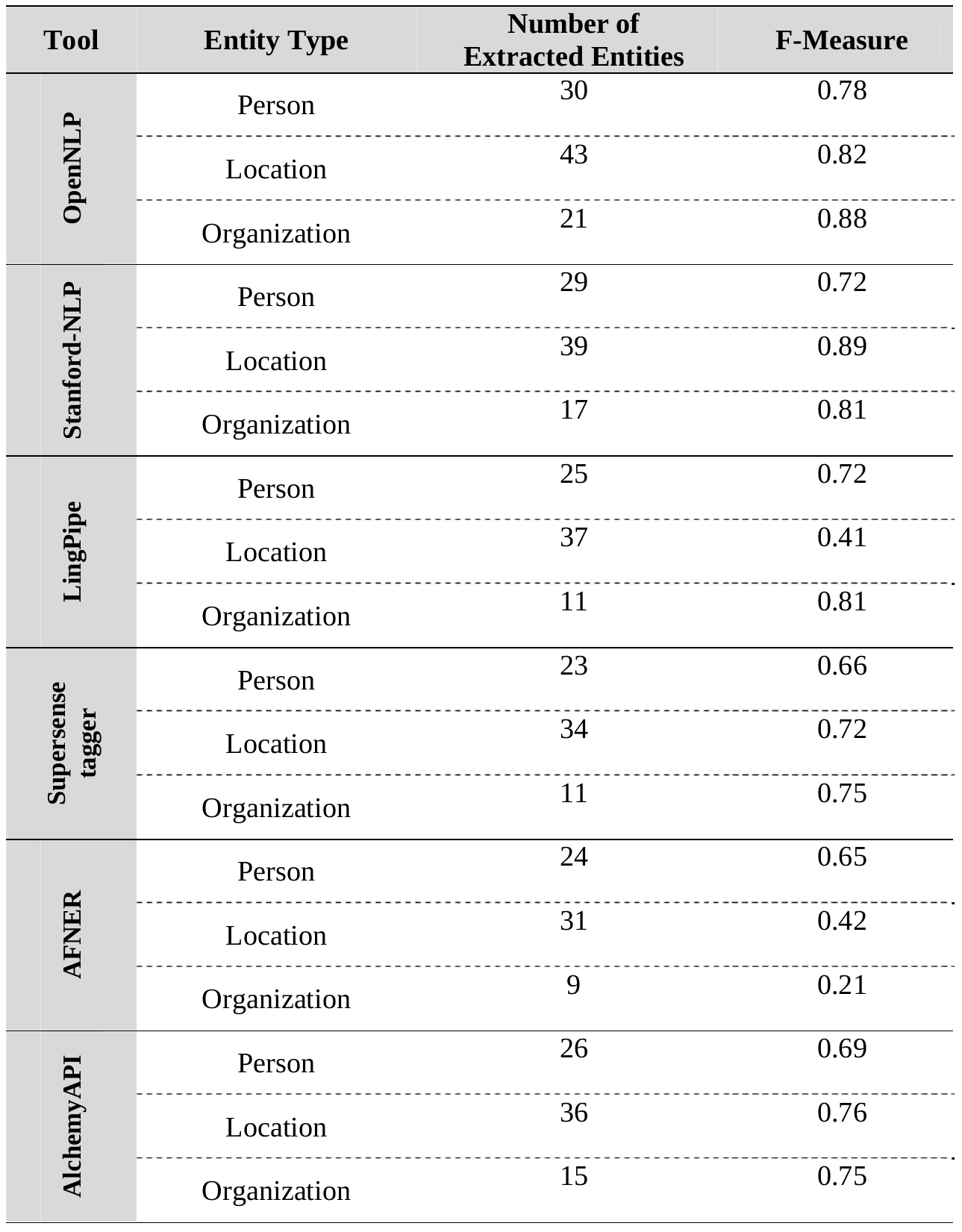}\\
  \end{tabular}
 \label{PRentityType}
\end{table}

We took a further analysis in account for the false positive errors, i.e. the elements erroneously identified as entities, as this could result more damaging in a project than partial identification or erroneous classification. To achieve this, in Table~\ref{PRentityType} we calculate the number of persons, locations and organizations and the f-measure for them. In this experiment, we didn't analyze the number of categories that each tool can recognize, as the utility and difficulty of recognition of some types against some others is different and demonstrates the need for a study based on the entity's types.

In this context, the study was carried out for person, location and organization types that the tools were able to recognize in the corpus. The analysis illustrated in Table~\ref{PRentityType} allows us to observe some differences to the global analysis. For example it is remarkable how OpenNLP has an f-measure on the entity type Person of 0.78, whilst AFNER achieves 0.65. As another example, Stanford-NLP has an f-measure on the entity type Location of 0.89, whilst LingPipe achieves~0.41.

\subsection{Tools for Entity Classification and their evaluation}

\begin{figure}[t]
\centering
\includegraphics[width=0.95\textwidth]{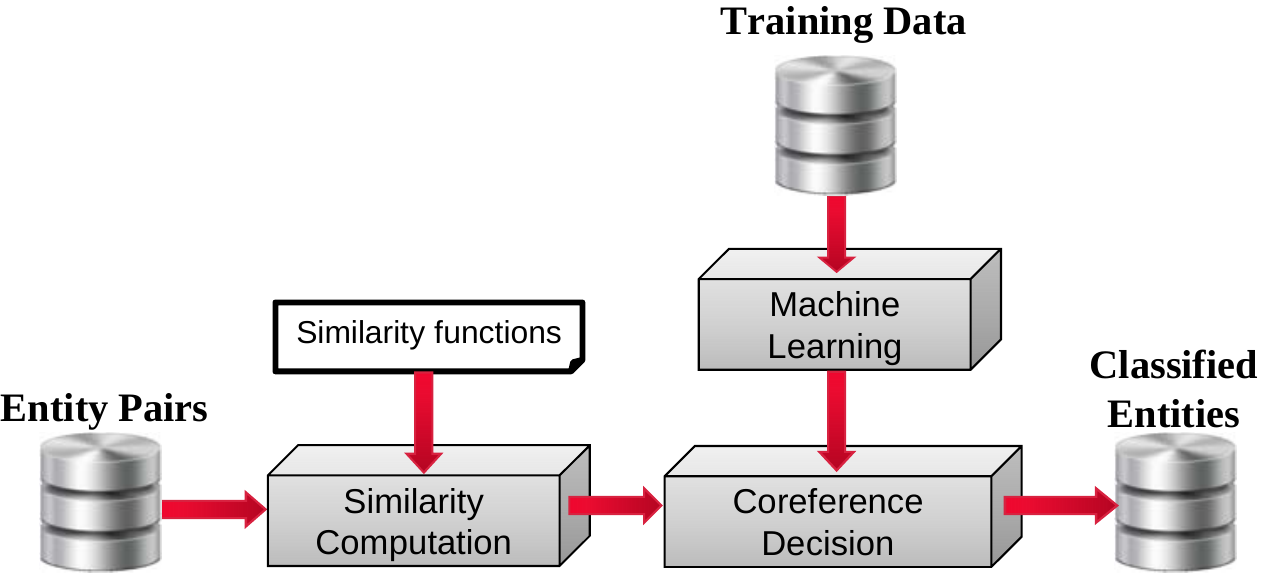}
\caption{Coreference classification process.}
\label{fig:classificationStep}
\end{figure}

The classification step compares pairs of entities, in which each entity is augmented with several features extracted from documents in the featurization step, and then determines whether these pairs of entities are coreferent or not. This step consists of two consecutive tasks (in Figure~\ref{fig:classificationStep}): \textit{similarity computation} and \textit{coreference decision}. The similarity computation task takes as input a pair of entities and computes the similarity scores between their features (e.g., character-, document-, or metadata-level features) using different appropriate similarity functions for the features. The coreference decision task classifies entity pairs as either ``coreferent'' or ``not coreferent'' based on the computed similarity scores between their features.

There are two alternative methods for the final coreference decision as follows: (i)~Threshold-based classification: The feature similarity scores of an entity pair might be combined by taking a weighted sum or a weight average of the scores. The entity pairs whose combined score is above a given threshold are considered as ``coreferent''; and (ii)~Machine learning-based classification: A classifier is trained by one of machine learning techniques (e.g., SVM or decision tree) using a training data and entity pairs are classified based on the trained classifier. The similarity scores between entity pairs are used as features for classification. For the similarity computation and the threshold-based coreference decision we use the following open-source packages:

\begin{itemize}
  \item \textbf{SecondString and SimMetrics: } \textit{SecondString}\footnote{http://secondstring.sourceforge.net} and \textit{SimMetrics}\footnote{http://sourceforge.net/projects/simmetrics/} are open-source packages that provide a variety of similarity functions used for comparing two feature attribute values. They provide different sets of similarity functions, e.g., \texttt{SecondString} does not provide \texttt{cosine} function supported by \texttt{SimMetrics}. Thus, we use both of the packages as we want to test different similarity functions for different cases.
  \item \textbf{Weka: } \textit{Weka}~\cite{Witten2005} is a free software package under the GNU public license, which is a collection of various machine learning algorithms developed in Java. It also provides functionalities for supporting some standard data mining tasks, such as data preprocessing, classification, clustering, regression and feature selection. The package can be applied in this project if a sufficient, suitable and balanced training data is available.
\end{itemize}

\subsubsection{Analysis and Methodology}

%

\begin{figure}[t]
\centering
\includegraphics[width=0.95\textwidth]{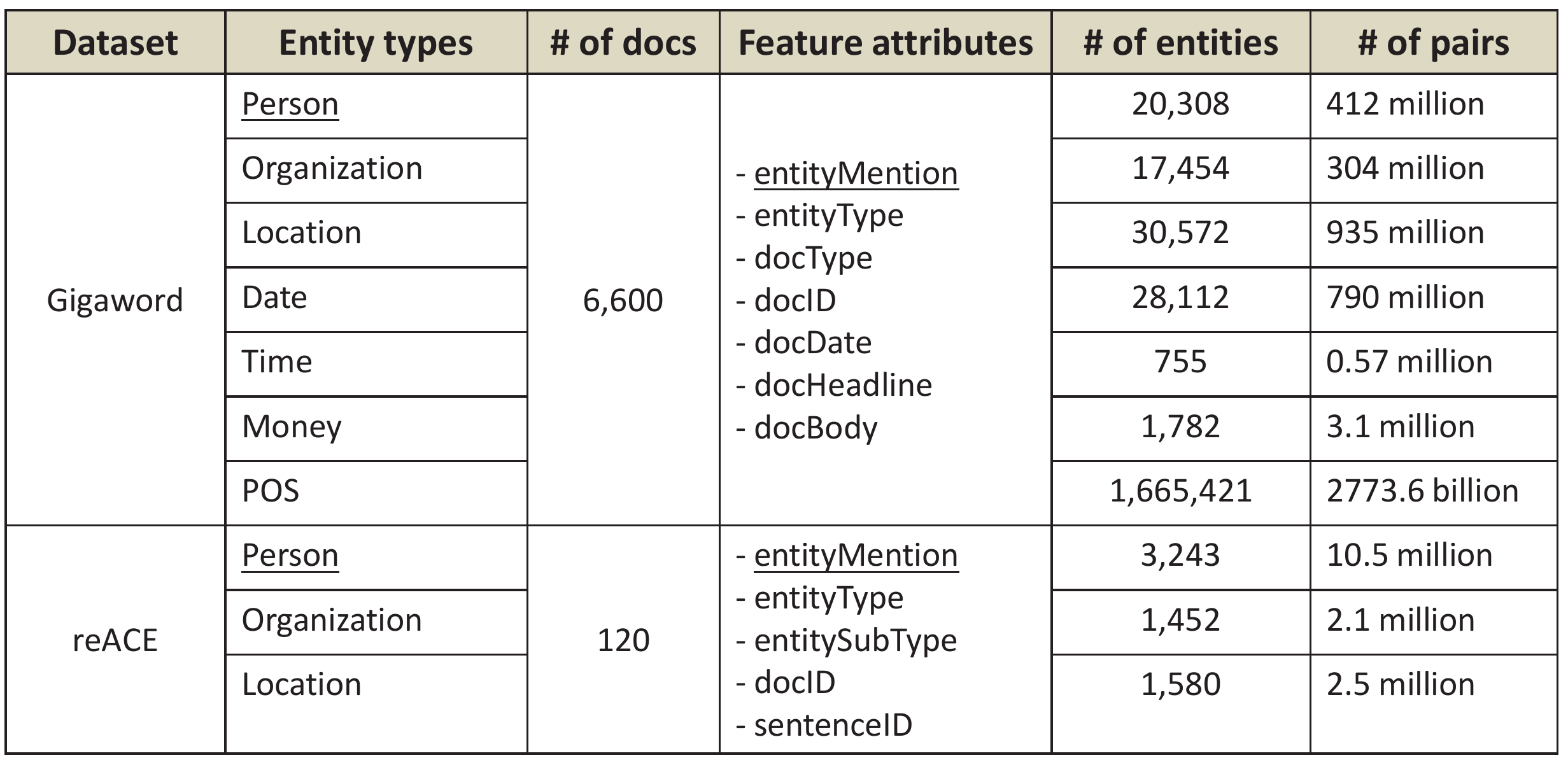}
\caption{Characteristics of datasets. The entity type and the feature attribute, which are considered in the evaluation, are underlined.}
\label{fig:dataset}
\end{figure}

\begin{figure}[t]
\centering
\includegraphics[width=1\textwidth]{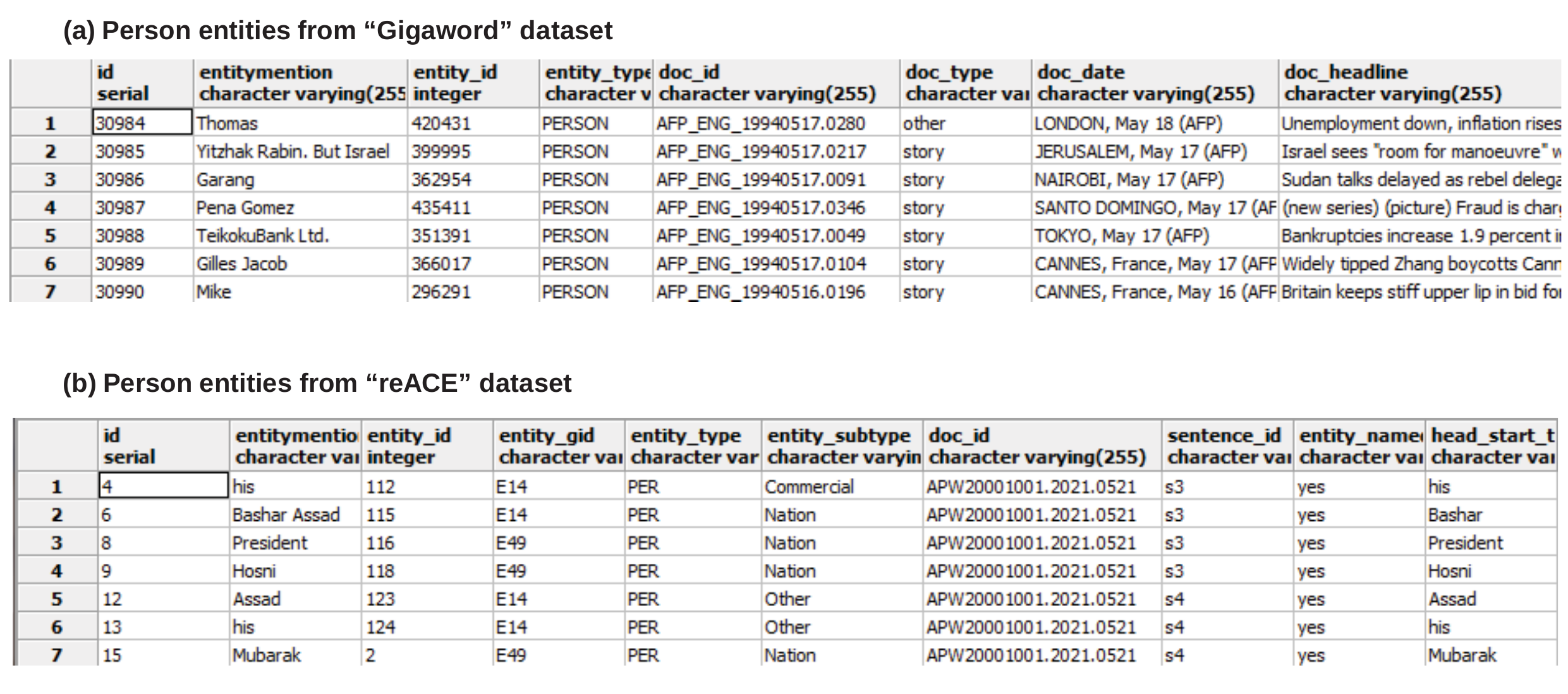}
\caption{Example person entities from two datasets.}
\label{fig:exampleEntities}
\end{figure}

In this assessment, we use reAce (to evaluate the effectiveness of the results) and English Gigaword (to evaluate the efficiency of the results) datasets. These datasets have been discussed in Section~\ref{chap1}. Figure~\ref{fig:dataset} shows the characteristics of those two datasets, which indicate for each dataset the types of extracted entities, the number of involved entities, the number of available feature attributes, the number of entity pairs, and so on. Figure~\ref{fig:exampleEntities} shows some example person entities (including metadata such as document identifier, type, and title) from the two datasets.

We measured the overall performance with both of \textit{efficiency} and \textit{effectiveness}. First, the efficiency is commonly determined in terms of the execution time which is taken in comparing feature attributes using similarity functions and then making coreference decisions based on their computed similarity scores. Second, the effectiveness is determined with the standard measures precision, recall and F-measure with respect to ``perfect coreference results'' which are manually determined. Let us assume that TP is the number of true positives, FP the number of false positives (wrong results), TN the number of true negatives, and FN the number of false negatives (missing results).

\begin{itemize}
	\item Precision= $\frac{TP}{TP+FP}$;
	\item Recall=  $\frac{TP}{TP+FN}$;
	\item F-measure=  $\frac{2*Precision*Recall}{Precision+Recall}$;
\end{itemize}

\begin{table}
 \centering
  \begin{tabular}{cc}
    \includegraphics[width=0.65\textwidth]{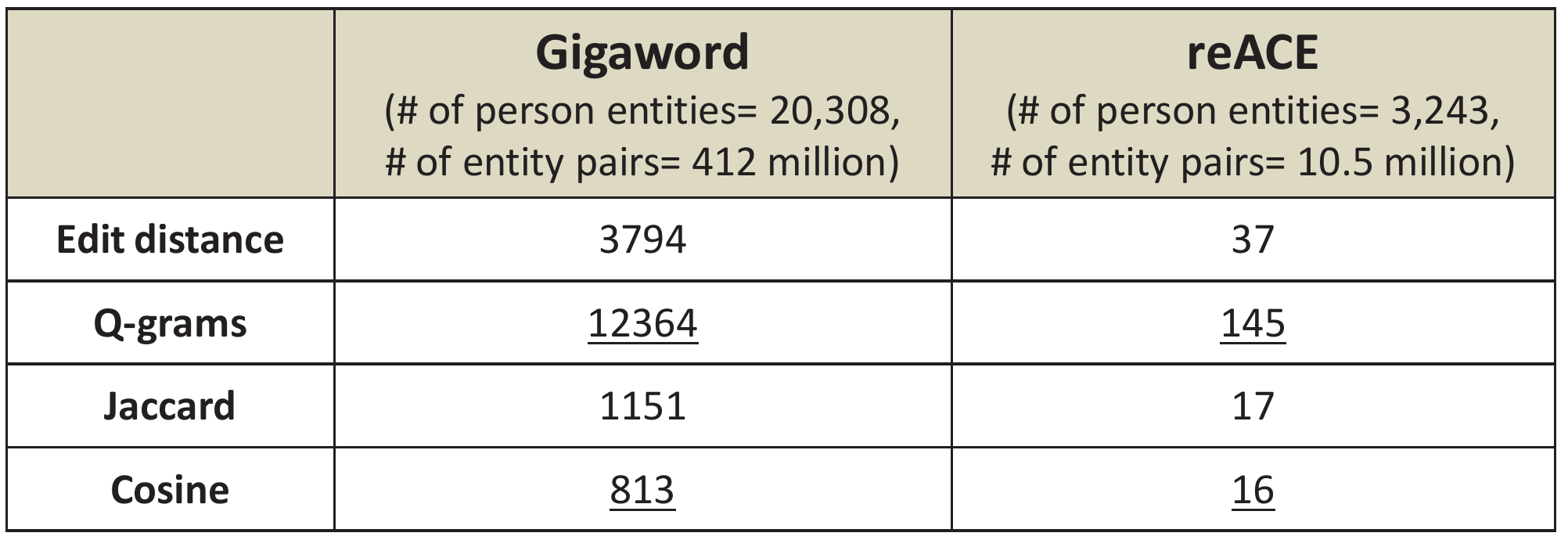}\\
  \end{tabular}
  \caption{Execution times (in seconds) for the two datasets. The smallest and largest values are underlined.}
 \label{tab:executionTime}
\end{table}

For the initial evaluations we focus on making coreference decisions for entity pairs of \texttt{Person} entity type. It should be noted that the same techniques described below would be applied to the other entity types, such as \texttt{organization}, \texttt{location}, and \texttt{date/time}. We compared feature attributes using various similarity functions. Figure~\ref{fig:dataset} indicates that several feature attributes could be used for the coreference decision (e.g., entity mention, docType, docDate, docHeadline, and docBody for the ``Gigaword'' dataset). In addition, we used the following four string similarity functions: \texttt{edit distance}, \texttt{Q-grams}, \texttt{jaccard}, and \texttt{cosine} functions. Here, edit distance and Q-grams are character-based functions while jaccard and cosine functions are token-based functions. For the \textit{Gigaword} dataset we only measured the \textit{execution time} as the perfect coreference results are not available. We applied the four similarity functions on one feature attribute (i.e., entity mention feature). For the \textit{reACE} dataset we measured the \textit{execution time} as well as the \textit{accuracy}. As in the ``Gigaword'' dataset, we used the four similarity functions in comparing the entity mention feature.  \\[-15pt]

\begin{figure}
\centering
\includegraphics[width=0.59\textwidth]{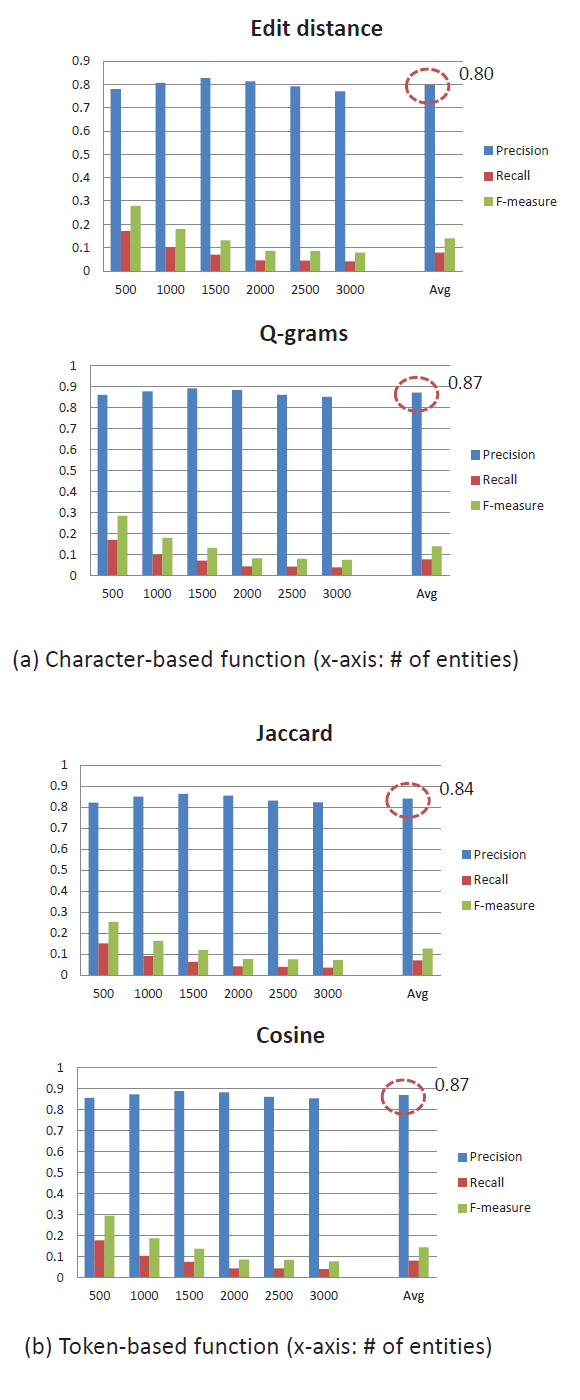}

\vspace{-3mm}
\caption{Evaluation results with the four different similarity functions (threshold= 0.5).}
\label{fig:precisionRecall}
\end{figure}

Table~\ref{tab:executionTime} lists the execution times taken for making coreference decisions by comparing \texttt{person} entities of the two datasets. The table shows significant differences between the applied similarity functions. The token-based functions (\texttt{Jaccard} and \texttt{cosine}) achieved fast execution time, when compared to the character-based functions (\texttt{edit distance} and \texttt{Q-grams}). This may be influenced by the algorithms of those functions, e.g., the character-based functions consider characters and their positions within strings to estimate the similarity, rather than considering tokens within strings as in the token-based functions. For the both datasets, among all the functions, the \texttt{Q-grams} function is the slowest one while the \texttt{cosine} function is the fastest one. 
When comparing 20,308 entities (the number of entity pairs is 412 millions) from ``Gigaword'' dataset, an execution time of 12,364 seconds is needed with the \texttt{Q-grams} function while an execution time of 813 seconds is needed with the \texttt{cosine} function.

Figure~\ref{fig:precisionRecall} shows the coreference quality (precision, recall, and F-measure) results for the ``reACE'' dataset with different similarity functions. 
The top half shows the results obtained by applying the character-based functions on just one single feature attribute of ``reACE'' dataset, namely \texttt{person name} entity mention. Among the character-based functions, the \texttt{Q-grams} function (average precision: 0.87) worked better than the \texttt{edit distance} function (average precision: 0.80). The bottom half shows the results achieved by applying the token-based functions on the same feature attribute. Among the token-based functions, the \texttt{cosine} function (average precision: 0.87) achieved slightly better results, compared with the \texttt{jaccard} function (average precision: 0.84). Among the functions, the coreference decision using the \texttt{Q-grams} function performed best while the one using the \texttt{edit distance} performed worst. The reason is that, if person names have multiple tokens and the tokens of the names are ordered differently in the other names, the \texttt{Q-grams} function could be more effective than the other character-based function, such as \texttt{edit distance}. All the functions performed reasonably well in terms of precisions, but they all suffered from very low recall, which means they missed many true coreferent entity pairs that should be contained in the returned results.

\section{Conclusions and Future Work}
\label{chap6}

In this paper we discussed the central concepts, subtasks, and the current state-of-the-art in Cross-Document Coreference Resolution (CDCR) process.
We provided assessment of existing tools/techniques for CDCR subtasks and highlight big-data challenges in each of them to help readers identify important and outstanding issues for further investigation. Finally, we provide concluding remarks and discuss possible directions for future work.
%
%
We believe that this is an important research area, which will attract a lot of attention in the research community.
In the following, we summarize significant research directions in this area.
\\\\
\textbf{Entity Extraction and the Big Data.} The entity extraction task outputs more data than it takes. Millions of documents can be used as an input to this task, and billions of entities can be extracted. In this context, the performance of entity extraction as well as the accuracy of extracted named entities should be optimized.
For example, as depicted in table~\ref{tab:executionTime} in Section~\ref{chap4}, the evaluation results on the effectiveness shows that the recall can be very poor, compared with the precision. There is a strong need for improving the recall results by exploiting more useful features and applying appropriate similarity functions to those features.
In this context, various load balancing techniques can be used to optimize the performance of MapReduce in extracting entities from huge number of documents. Moreover, various dictionaries and knowledge bases such as YAGO, freebase, DBpedia, and reACE can be used for training which may help to optimize the accuracy of extracted entities.
\\\\
\textbf{Entity Pairs Filtering and Featurization of Billions Extracted Entities.} For huge number of extracted entities, it is generally not feasible to exhaustively evaluate the Cartesian product of all input entities, and generate all possible entity pairs. To address this challenge, various blocking techniques (e.g., blocking strategy for all non-learning and learning-based match approaches) can be used to reduce the search space to the most likely matching entity pairs. Moreover, featurization of the corpus as well as extracted entities, will facilitate the filtering step and also will quickly eliminates those pairs that have little chance of being deemed co-referent. Similar to entity extraction phase, generating a knowledge-base from existing Linked Data systems may facilitate the featurization~step.
\\\\
\textbf{Classification of Billions Entity Pairs.} Various machine learning over a set of training examples can be used to classify the pairs as either co-referent or not co-referent. Different approaches has different similarity threshold, where entity pairs with a similarity above the upper classification threshold are classified as matches, pairs with a combined value below the lower threshold are classified as non-matches, and those entity pairs that have a matching weight between the two classification thresholds are classified as possible matches. This task is challenging as we need to investigate how different configurations could have an impact on the effectiveness and efficiency of coreference classification. Three characteristics can be considered for this configuration: (i)~which feature attributes to be used for classification; (ii)~which similarity functions to be used for the chosen feature attributes; and (iii)~which threshold is suitable for the classification decision.
\\\\
\textbf{Clustering of Billions of (Cross Document) Co-referent Entities.}
Once the individual pairs are classified, they must be clustered to ensure that all mentions of the same entity are placed in the same equivalence class.
Standard entity clustering systems commonly rely on mention (string) matching, syntactic features, and linguistic resources like English WordNet. Challenges here include: (i)~assigning each cluster to a global entity. For example, the cluster including ``Obama, B. Obama, B.H. Obama, Barack Obama, Barack H. Obam, etc" should be considered as mentions of the global entity `President of the United State'. To achieve, Linked Data systems can be used to help identifying the entities; and (ii)~when co-referent text mentions appear in different languages, standard entity clustering  techniques cannot be easily applied.

\newpage

\bibliographystyle{plain}
\bibliography{sample}

\newpage

\section*{Appendix: MapReduce-based Software Prototype}
\label{Appendix}

\begin{figure}
\begin{adjustwidth}{-1cm}{}
\centering
\includegraphics[width=0.8\textwidth]{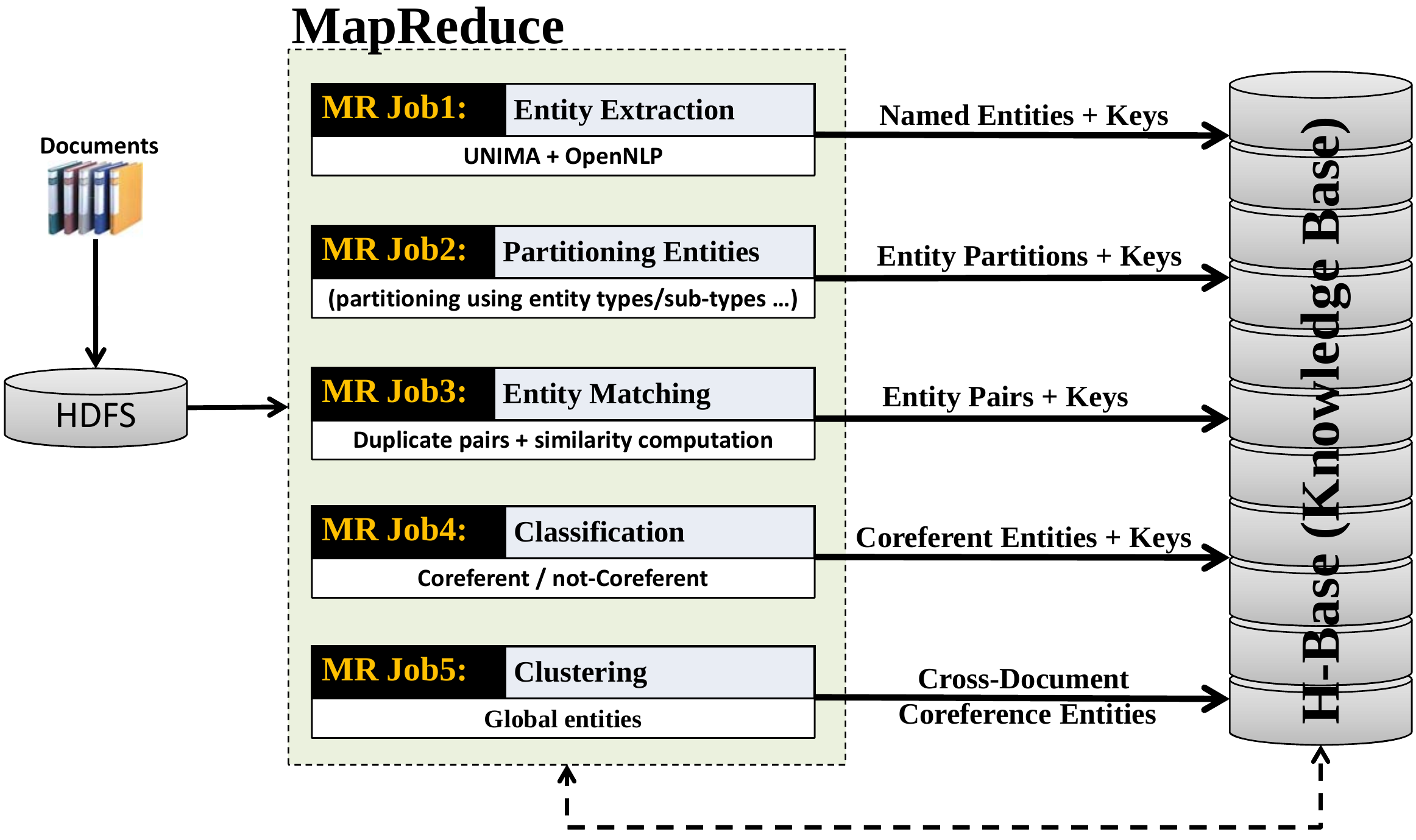}
\caption{CDCR MapReduce Process.}
\label{fig:MapReduceProcessGeneral}
\end{adjustwidth}
\end{figure}

\begin{figure}
\begin{adjustwidth}{-1cm}{}
\centering
\includegraphics[width=1.1\textwidth]{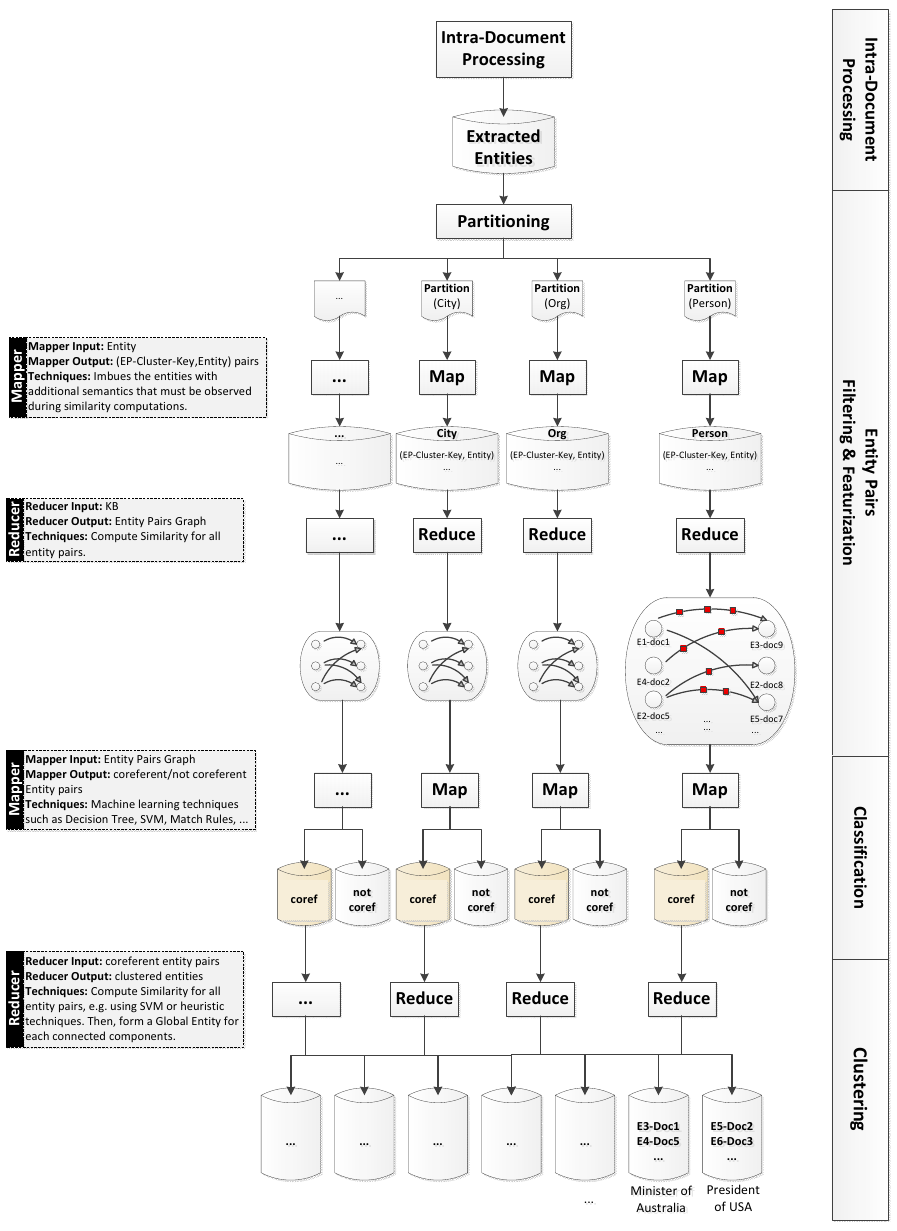}
\caption{CDCR MapReduce Process.}
\label{fig:MapReduceProcess}
\end{adjustwidth}
\end{figure}

To address the big data challenges discussed in this paper, we have developed a MapReduce-based software prototype to identify huge number of entity mentions (e.g., persons, organizations or locations) across huge amount of multiple documents that refer to the same underlying entity, i.e., the process of cross-document coreference resolution (CDCR). Figures~\ref{fig:MapReduceProcessGeneral} and ~\ref{fig:MapReduceProcess} illustrate the overall MapReduce architecture for this process. As illustrated in this figure, specified workflows are automatically translated into MapReduce jobs for parallel execution on different Hadoop clusters. In particular, five MapReduce jobs are specified: Entity Extraction, Entity Partitioning, Entity Matching, Classification, and Clustering.

\begin{figure} [hb]
\centering
\includegraphics[width=0.65\textwidth]{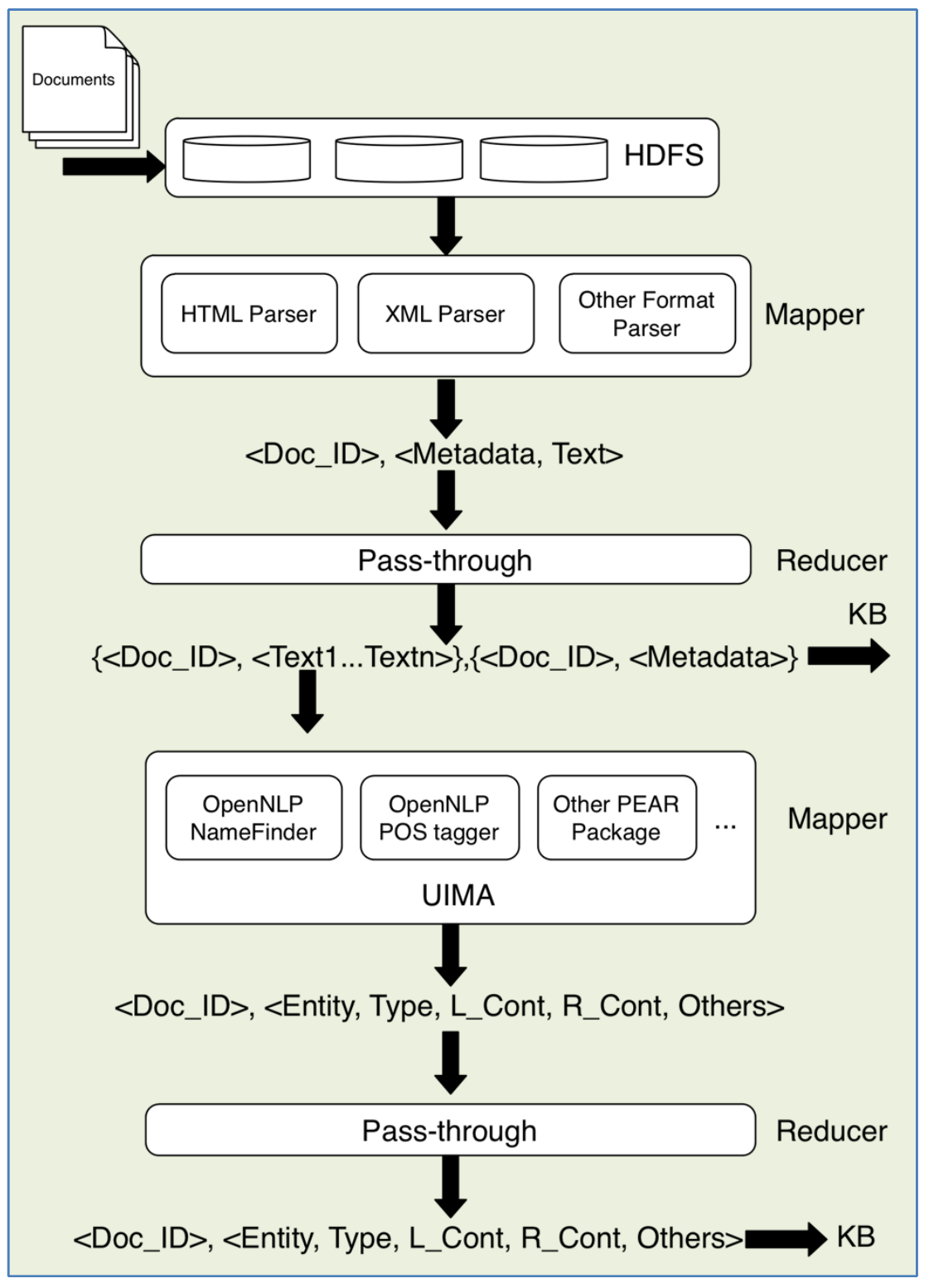}
\caption{Entity extraction task in the MapReduce-based software prototype.}
\label{fig:MapReduceProcessEntity}
\end{figure}

\begin{figure}
\begin{adjustwidth}{-1cm}{}
\centering
\includegraphics[width=0.8\textwidth]{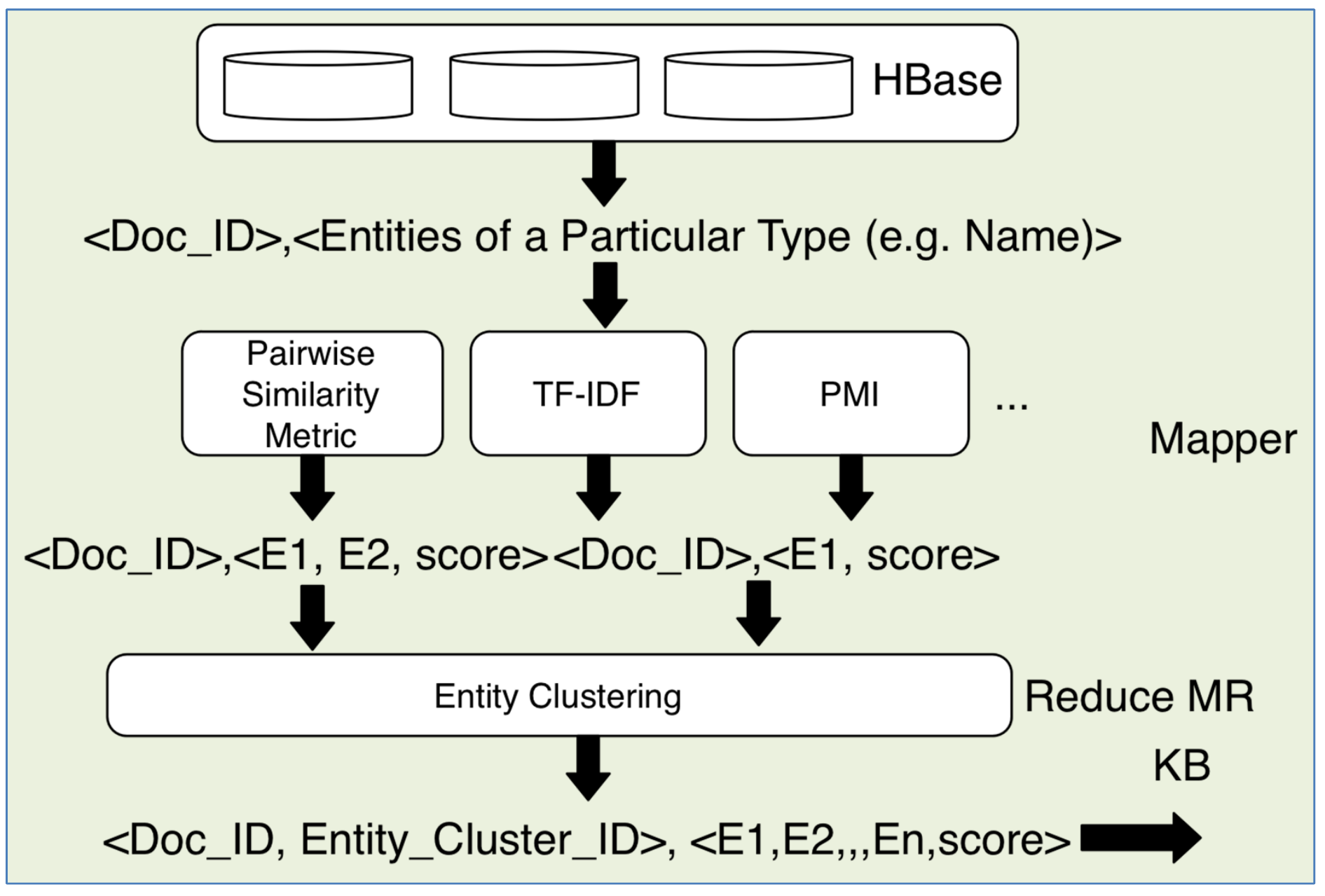}
\caption{Entity clustering task in the MapReduce-based software prototype.}
\label{fig:MapReduceProcessCluster}
\end{adjustwidth}
\end{figure}

\begin{itemize}
  \item \textbf{MR Job1:} Entity Extraction. In this phase we use UNIMA and OpenNLP to extract named entities. Figure~\ref{fig:MapReduceProcessEntity} illustrates details of this MapReduce job, where set of documents will be feed as input into the Hadoop File System (HDFS). Afterward, set of mappers will prepare documents for tokenization. Finally, using UIMA and OpenNLP, set of extracted named entities and some related metadata such as the document ID which the entity has been extracted from, the type/sub-type of the entity, and document timestamp will be generated as the output of the first MapReduce phase. Entities and associated attributes will be stored into distributed database (Apache Hbase\footnote{http://hbase.apache.org/}, i.e., the Hadoop database which is a distributed scalable big data store) and can be considered as the knowledge base (KB).
  \item \textbf{MR Job2:} Entity Partitioning. As mentioned earlier, pairwise entity comparison can become exponential and very time consuming. To avoid this, we partition named entities according to their type and subtype. For example, all entities typed as `person' and sub-typed as `politician' will be stored in the same partition. Each partition will be send to different (MapReduce) mappers to generate candidate entity pairs.
  \item \textbf{MR Job3:} Entity Matching. After partitioning the entities, the third MapRedcue job will be responsible for pairing entities in each constructed partition. Figure~\ref{fig:MapReduceProcessCluster} illustrates details of this MapReduce job, where duplicate pairs will be identified, the importance of the selected entities to a document in the corpus will be identified, and pairwise similarity degree will be calculated for each pair. As a result set of candidate pair entities will be generated and feed into next MapReduce job.
  \item \textbf{MR Job4:} Classification. Candidate pair entities generated in previous MapReduce job, will be feed to this MapReduce job, where entity pairs with  similarity degree (i.e., a real number between 0 and 1) more than, or equal to, 0.5 will be considered as \emph{coreferent} otherwise they will be considered as \emph{not-coreferent}.
  \item \textbf{MR Job5:} Clustering. Set of coreferent entities generated in previous MapReduce job, will be feed into this MapReduce job. Figure~\ref{fig:MapReduceProcess} illustrates more details of these steps and the MapReduce mappers and reducers. In particular, set of supervised and/or unsupervised algorithms can be used to analyze coreferent entities, recognize patterns among them, and classify them in different clusters. We used decision tree like algorithm (see Section~\ref{chap4} for details) to classify coreferent entities.
\end{itemize}

\subsection{Software Prototype Evaluation}

We evaluated the performance of the software prototype on a cloud system having: (i)~One Head Node, having one QuadCore 2.33 GHz processor, 8 GB RAM, 140 GB storage, and with 15TB RAID6 disk array attached; and (ii)~Two Identical Blade Servers, having four QuadCore 2.4 GHz processor, 96 GB ram, 1.2 TB storage which configured as a private cloud using OpenStack. We configured the cloud using Apache Hadoop 1.0.4 and ran MapReduce CDCR over clusters of 1, 2 and 4 4-core virtual-machines.

\begin{figure} [t]
\centering
  \includegraphics[scale=0.8]{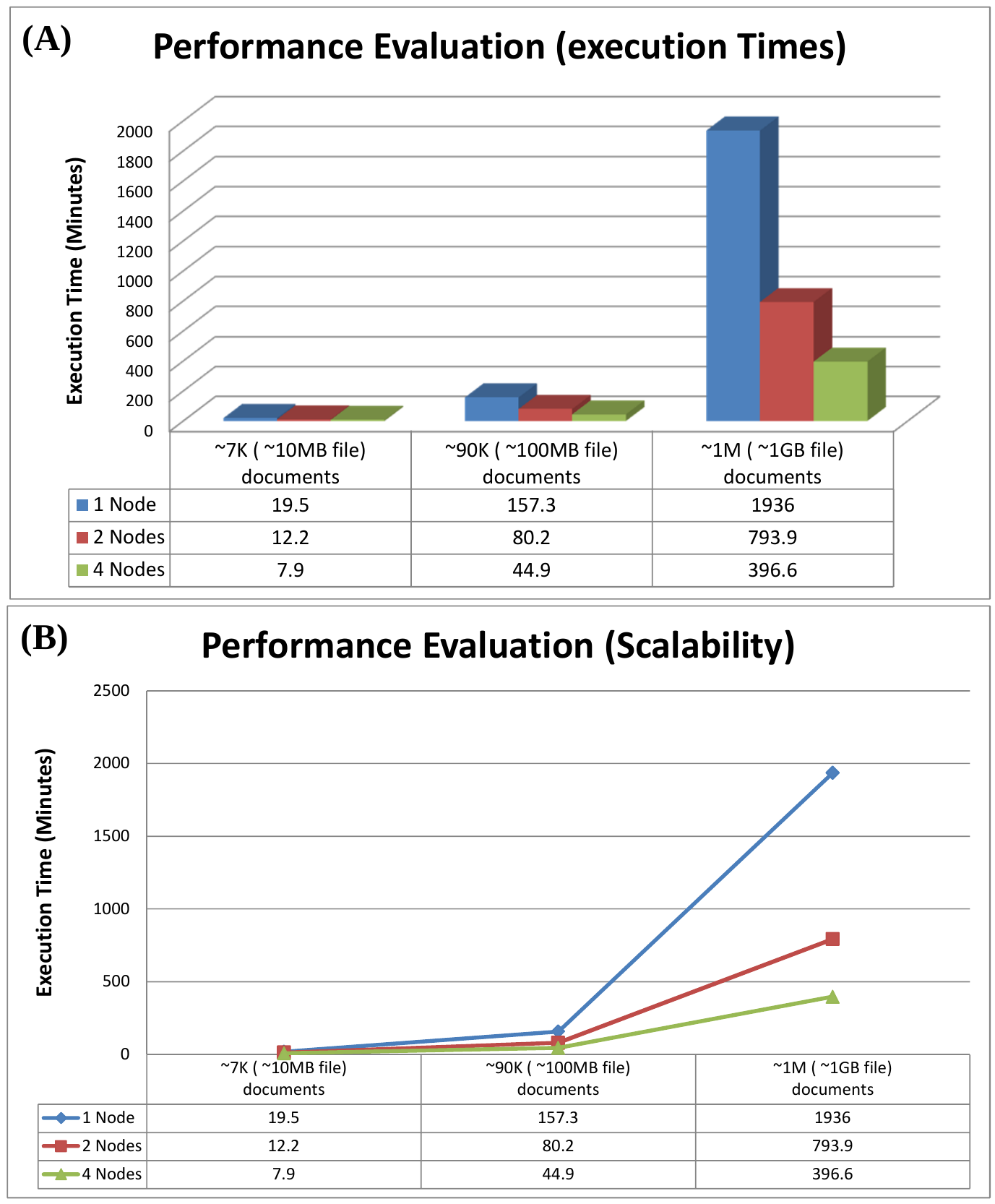}\\
  \caption{Software prototype evaluation: (A) execution times; and (B) Scalability.}\label{analytic0}
\end{figure}

We used Gigaword dataset (see Table~\ref{datasets}) for evaluating the software prototype. Figure~\ref{analytic0} illustrates the execution times and the scalability evaluation. As depicted in the figure, we divided each dataset into regular number of documents and ran the experiment for each of them. The evaluation shows the viability and efficiency of using MapReduce in cross-document coreference resolution process. Table~\ref{analytic1} illustrates the result of experiments over Gigaword dataset including samples of different number of documents, number of extracted entites from each sample, number of entity pairs for each group of extracted entities, and number of coreferent clusters for classified coreferent entities. Table~\ref{analytic2} illustrates a sample of extracted named entities from Gigaword dataset including the entity type, document ID which the entity has been extracted from, and some metadata about the document such as type of the document (e.g., sport and history) and document timestamp. Table~\ref{analytic3} illustrates a sample of paired entities and their similarity degree. And Table~\ref{analytic4} illustrates a sample of coreferent clusters generated from the coreferent entities.

\begin{table} [b]
 \caption{The result of experiments over Gigaword dataset including samples of different number of documents, number of extracted entities from each sample, number of entity pairs for each group of extracted entities, and number of coreferent clusters for classified coreferent entities.}
 \centering
  \begin{tabular}{cc}
   \includegraphics[scale=0.9]{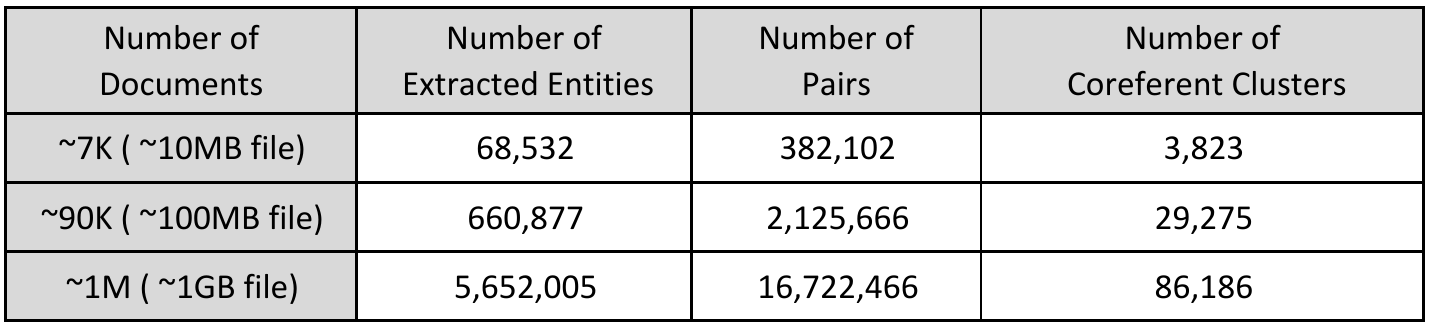}\\
  \end{tabular}
 \label{analytic1}
\end{table}



Table~\ref{analytic2} illustrates a sample of extracted named entities from Gigaword dataset including the entity type, document ID which the entity has been extracted from, and some metadata about the document such as type of the document (e.g., sport and history) and document timestamp. Table~\ref{analytic3} illustrates a sample of paired entities and their similarity degree. And Table~\ref{analytic4} illustrates a sample of coreferent clusters generated from the coreferent entities.
\\\\\\\\\\
As an ongoing work, we plan to use a graph-based clustering approach in the second mapper in Figure~\ref{fig:MapReduceProcessGeneral}. We will use our previous work, i.e. a Map-Reduce enabled graph processing engine~\cite{DBLP:conf/bpm/BeheshtiBNS11,DBLP:conf/wise/BeheshtiBNA12,DBLP:conf/caise/BeheshtiBN13}, to model the entities and the relationships among them as graphs. We will use On-Line Analytical Processing on Graphs~\cite{DBLP:conf/wise/BeheshtiBNA12} to create new relationship between entities mentions by calculating the similarity between them. We plan to employ the weighted support vectors and find the shortest path between two mentions in the graph to facilitate the classification and clustering of entities and their mentions across documents. Moreover, we plan to leverage crowdsourcing techniques~\cite{compton2009incremental,DBLP:conf/apweb/AllahbakhshIBBBF13,DBLP:conf/colcom/AllahbakhshIBBBF12} for incremental and end-user-centered knowledge acquisition to improve the classification of paired entities.

\begin{table} [b]
 \caption{Sample of coreferent clusters generated from coreferent entities in the MapReduce-based software prototype.}
 \centering
  \begin{tabular}{cc}
   \includegraphics[scale=0.8]{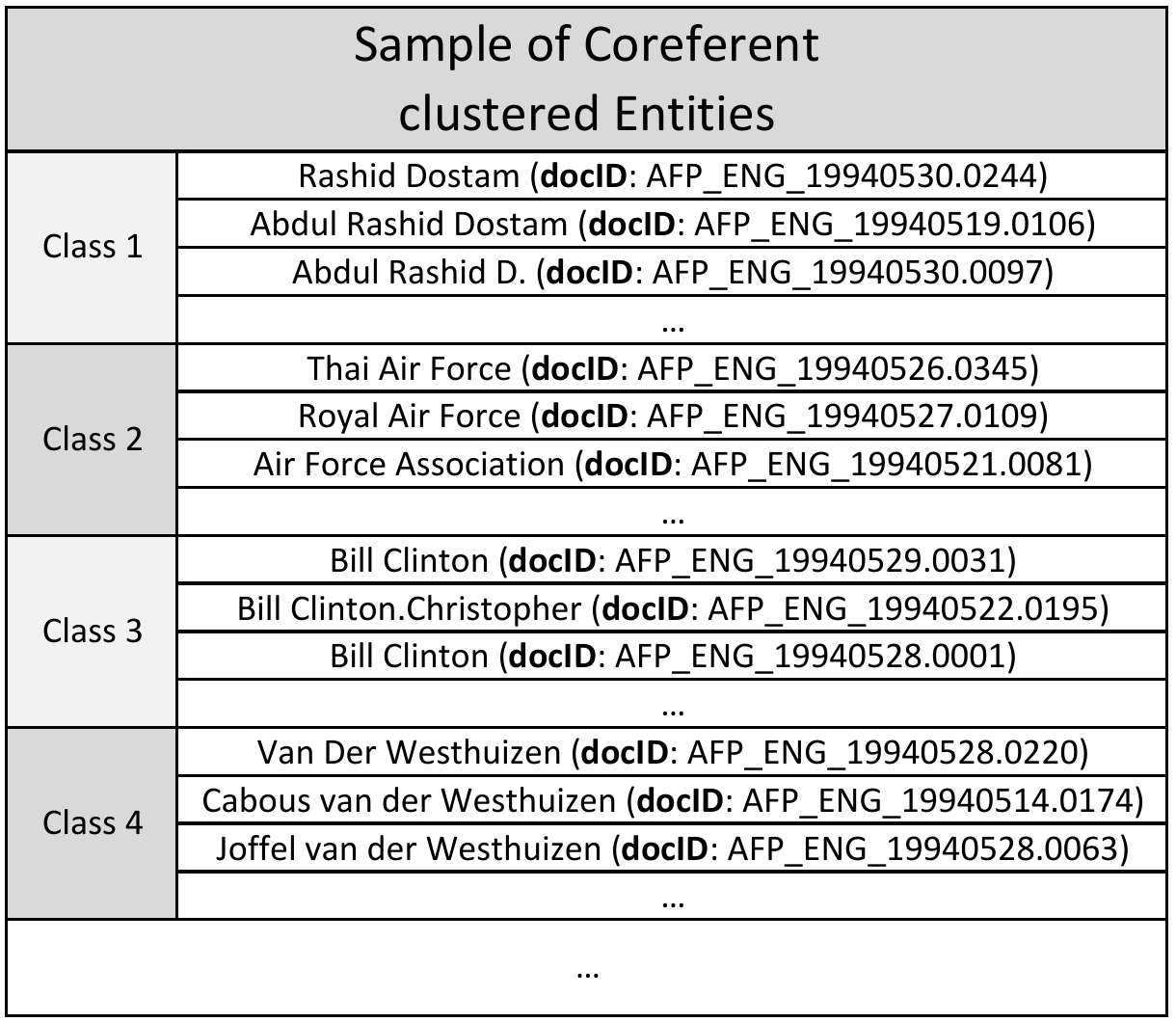}\\
  \end{tabular}
 \label{analytic4}
\end{table}

\begin{landscape}

\begin{table}
 \caption{Sample of extracted named entities from Gigaword dataset including the entity type, document ID which the entity has been extracted from, and some metadata about the document such as type of the document (e.g., sport and history) and document timestamp in the MapReduce-based software prototype.}
 \centering
  \begin{tabular}{cc}
   \includegraphics[scale=1.2]{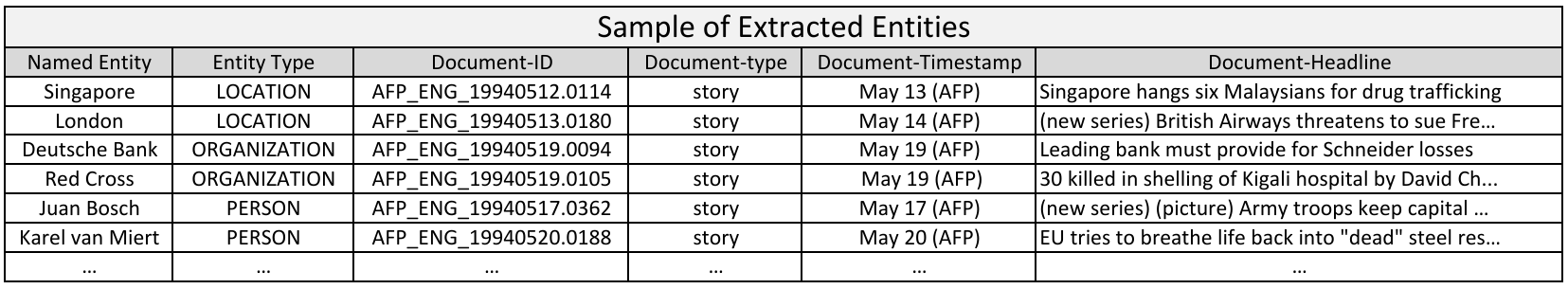}\\
  \end{tabular}
 \label{analytic2}
\end{table}

\begin{table}
 \caption{Sample of paired entities and their similarity degree in the MapReduce-based software prototype.}
 \centering
  \begin{tabular}{cc}
   \includegraphics[scale=1.2]{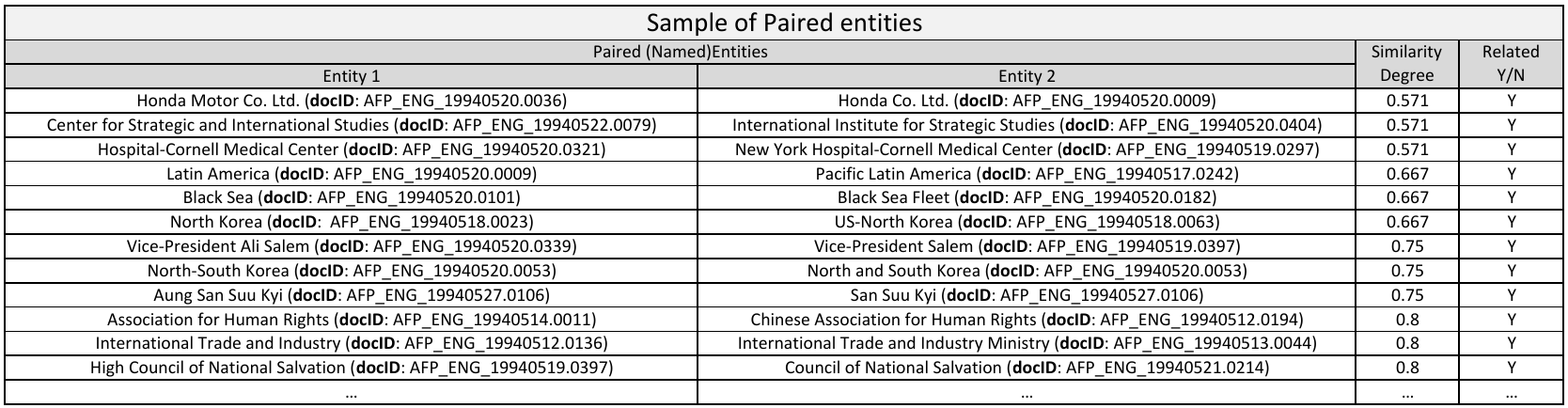}\\
  \end{tabular}
 \label{analytic3}
\end{table}

\end{landscape}

\end{document}